\documentclass[sigconf]{acmart}
\usepackage{algorithm}
\usepackage{algpseudocode}
\usepackage{graphicx}
\usepackage{subfig}
\usepackage{caption,acmart-taps}
\usepackage{multicol, multirow}
\usepackage{enumitem}
\AtBeginDocument{%
  \providecommand\BibTeX{{%
    \normalfont B\kern-0.5em{\scshape i\kern-0.25em b}\kern-0.8em\TeX}}}

\copyrightyear{2023}
\acmYear{2023}
\setcopyright{acmlicensed}\acmConference[DAI '23]{The Fifth International Conference on Distributed Artificial Intelligence}{November 30-December 3, 2023}{Singapore, Singapore}
\acmBooktitle{The Fifth International Conference on Distributed Artificial Intelligence (DAI '23), November 30-December 3, 2023, Singapore, Singapore}
\acmPrice{15.00}
\acmDOI{10.1145/3627676.3627685}
\acmISBN{979-8-4007-0848-0/23/11}
\newcommand{\minisection}[1]{\vspace{5pt}\noindent\textbf{#1.}}

\begin{document}

\title{ROMO: Retrieval-enhanced Offline Model-based Optimization}


\author{Mingcheng Chen}
\affiliation{%
  \institution{Shanghai Jiao Tong University}
  \city{Shanghai}
  \country{China}}
\email{mcchen@apex.sjtu.edu.cn}

\author{Haoran Zhao}
\affiliation{%
  \institution{Shanghai Jiao Tong University}
  \city{Shanghai}
  \country{China}}
\email{hrzhao@apex.sjtu.edu.cn}

\author{Yuxiang Zhao}
\affiliation{%
    \institution{China Mobile Research Institute}
    \city{Beijing}
    \country{China}}
\email{zhaoyuxiang@chinamobile.com}

\author{Hulei Fan}
\affiliation{%
    \institution{China Mobile (Zhejiang) Research \& Innovation Institute}
    \city{Hangzhou}
    \country{China}}
\email{fanhulei@zj.chinamobile.com}

\author{Hongqiao Gao}
    \affiliation{%
    \institution{China Mobile (Zhejiang) Research \& Innovation Institute}
    \city{Hangzhou}
    \country{China}}
\email{gaohongqiao@zj.chinamobile.com}

\author{Yong Yu}
    \affiliation{%
    \institution{Shanghai Jiao Tong University}
    \city{Shanghai}
    \country{China}}
\email{yyu@apex.sjtu.edu.cn}

\author{Zheng Tian}
\authornote{The corresponding author}
    \affiliation{%
    \institution{ShanghaiTech University}
    \city{Shanghai}
    \country{China}}
\email{tianzheng@shanghaitech.edu.cn}

\renewcommand{\shortauthors}{M. Chen, et al.}

\begin{abstract}
  Data-driven black-box model-based optimization (MBO) problems arise in a great number of practical application scenarios, where the goal is to find a design over the whole space maximizing a black-box target function based on a static offline dataset.
In this work, we consider a more general but challenging MBO setting, named constrained MBO (CoMBO), where only part of the design space can be optimized while the rest is constrained by the environment. A new challenge arising from CoMBO is that most observed designs that satisfy the constraints are mediocre in evaluation. Therefore, we focus on optimizing these mediocre designs in the offline dataset while maintaining the given constraints rather than further boosting the best observed design in the traditional MBO setting.
We propose \emph{retrieval-enhanced offline model-based optimization} (ROMO), a new derivable forward approach that retrieves the offline dataset and aggregates relevant samples to provide a trusted prediction, and use it for gradient-based optimization.
ROMO is simple to implement and outperforms state-of-the-art approaches in the CoMBO setting. Empirically, we conduct experiments on a synthetic Hartmann (3D) function dataset, an industrial CIO dataset, and a suite of modified tasks in the Design-Bench benchmark.
Results show that ROMO performs well in a wide range of constrained optimization tasks.
\end{abstract}


\maketitle

\section{Introduction}
Data-driven black-box model-based optimization, also known as offline model-based optimization (MBO), is widely involved in real-world domains, such as designing drug molecule \cite{ChEMBL} and protein sequence \cite{CbAS}; optimizing aircraft designs \cite{hoburg2014geometric}, robot morphologies \cite{liao2019data}, and controller parameters \cite{berkenkamp2016safe}; searching neural architectures \cite{zoph2017neural}; and finetuning industrial systems \cite{zhan2022deepthermal}.
These practical application scenarios require optimizing inputs $\mathbf{x} \in \mathcal{X}$ of a black-box function $f(\mathbf{x})$ whose evaluation for a $y$ is expensive or inaccessible, so most existing works consider making use of only static offline logged data $\mathcal{D}$ to optimize such a target function.

We consider constrained model-based optimization (CoMBO), a more general\footnote{The standard offline MBO can be included as a special case of CoMBO when $\tilde{\mathcal{X}} = \mathcal{X}$.} offline setting where some input dimensions are fixed throughout the optimization process, so optimizations are only allowed over a subspace of the whole design space $\tilde{\mathcal{X}} \subset \mathcal{X}$.
In CoMBO, offline samples satisfying constraints often exhibit far-from-optimal performance, or samples satisfying the constraints are rare or absent in a high-dimensional design space.
Therefore, we cannot guarantee obtaining a candidate solution that is no worse than the best observation by simply assigning the best solutions from the offline dataset $\{\textbf{x}^*_\mathcal{D} | \forall \textbf{x}\in\mathcal{D}, f(\textbf{x})\leq f(\textbf{x}^*_\mathcal{D})\}$ to the target design in this setting.
Accordingly, we study how to optimize those mediocre designs in the offline dataset toward the optimal while satisfying the given dimensional constraints in the CoMBO setting.

\begin{figure}[tbp]
    \centering
    \includegraphics[width=0.985\linewidth]{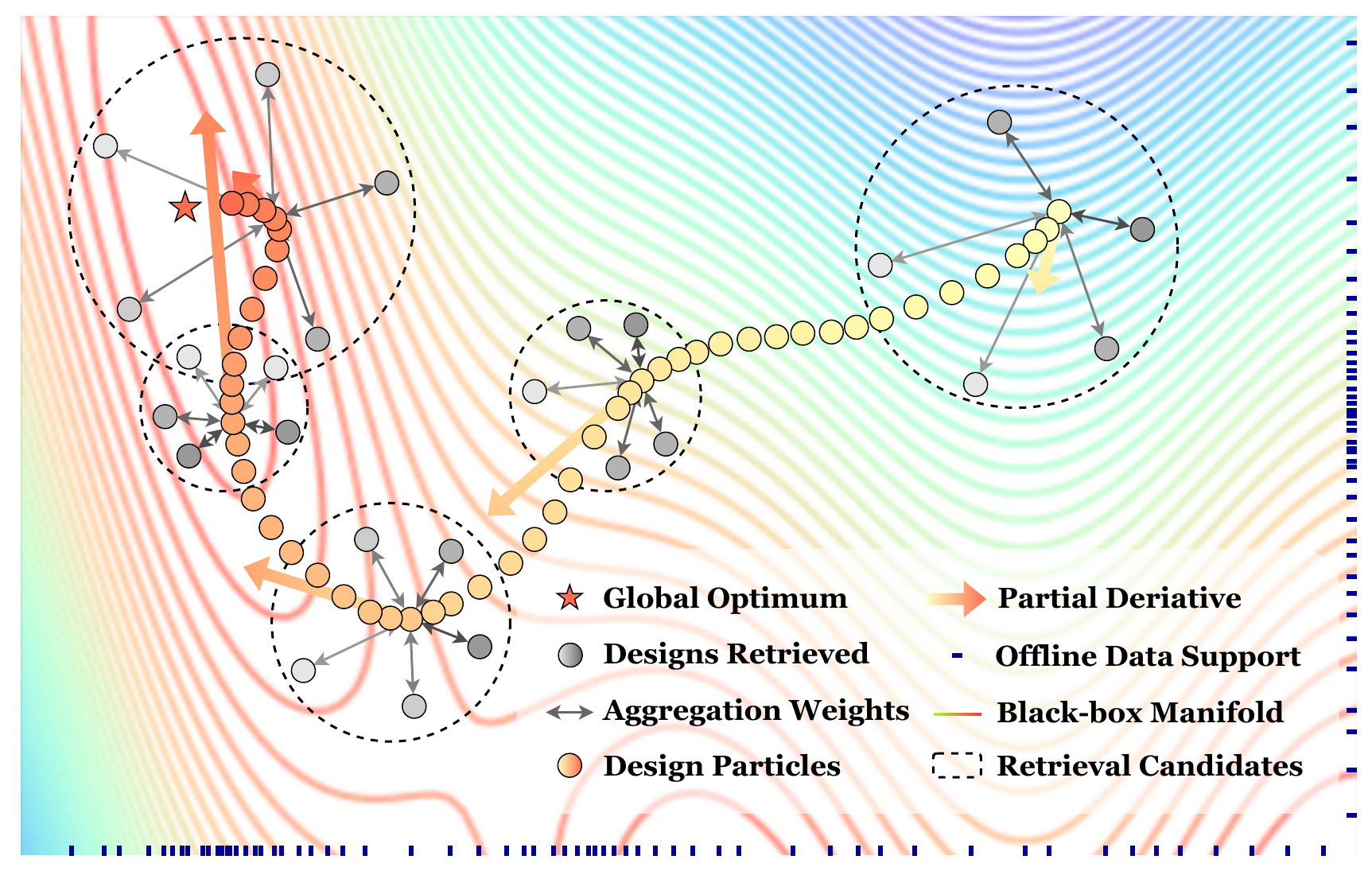}
    \caption{Illustration showing the intuition of ROMO on a 2D modifiable design subspace.}
    \label{fig:intuition}
    \vspace{-10pt}
\end{figure}

Hence, in this work, we propose a new derivable surrogate modeling approach called \emph{Retrieval-Enhanced Offline Model-Based Optimization} (ROMO) to handle the proposed CoMBO problem (Figure~\ref{fig:intuition}).
Given mediocre designs satisfying certain dimensional constraints, ROMO first constructs a set of retrieval candidates for the current forward approach by ranking the similarity between designs.
Then a retrieval-enhanced forward network aggregates retrieved designs and corresponding scores to make a reliable reference prediction.
To eliminate the additional dimensions introduced by aggregations that affect the gradient-based boosting process, the retrieval-enhanced forward model is subsequently distilled onto another surrogate forward network.
The gradients thus will be primarily dominated by those dimensions that are allowed to be modified on this lower-dimensional surrogate network.
At the same time, additional regularizations are also incorporated into the model training process to ensure that the surrogate network produces effective predictions while avoiding overestimations as possible.

The contributions of this work are summarized as follows:
\begin{itemize}[leftmargin=12pt]
    \item We raise a novel and important MBO problem, i.e., constrained model-based optimization (CoMBO), which is a more general offline setting and is common in real-world domains. 
    \item We analyze the performance of state-of-the-art derivable surrogate approaches under such a CoMBO setting, revealing the problem of over-conservatism and demonstrating how the defect affects the optimization process when starting from a low-score design.
    \item We propose retrieval-enhanced offline model-based optimization (ROMO), a simple but effective forward surrogate-based modeling approach. It employs a retrieval-enhanced forward network to provide conservative but reliable reference prediction for the inference of the surrogate network. To our knowledge, ROMO is the first approach that introduces a retrieval mechanism into offline model-based optimization, and it works well in both real-world CoMBO task and modified offline MBO benchmark.
\end{itemize}

We test ROMO on a Hartmann (3D) test function to explain why ROMO can perform better than prior work in the CoMBO setting and reveal the defects of existing offline MBO approaches.
Furthermore, experiments on a suite of modified tasks in Design-Bench \cite{design-bench} and a real-world CIO load balancing task are also provided to validate the performance of ROMO under the CoMBO setting.

\section{Preliminaries}
\subsection{Problem Statement}
The goal of black-box optimization (MBO) is to find optimal solutions $\textbf{x}^*$ that maximize an unknown black-box function $f: \mathcal{X} \rightarrow \mathbb{R}$, where the design domain is denoted as $\mathcal{X} := \{\textbf{x} | \textbf{x} \in \mathbb{R}^d\}$:
\begin{equation}
    \textbf{x}^* \leftarrow \mathop{\arg\max}_{\textbf{x}\in\mathcal{X}} f(\textbf{x}).
\end{equation}

In the setting of \emph{data-driven} black-box model-based optimization, also known as \emph{offline} MBO, the evaluations of $f(\textbf{x})$ for any $\textbf{x}$ are not allowed at training time.
Instead of directly querying the ground-truth black-box function $f$, offline approaches must utilize an offline static dataset $\mathcal{D}$ of input designs and their objective scores, $\mathcal{D} = \{(\textbf{x}_1,y_1),(\textbf{x}_2,y_2),\dots,(\textbf{x}_N,y_N)\}$, where $y_t = f(\textbf{x}_t)$.
Since directly locating the optima is very difficult, typically, we are allowed a small budget of $Q$ queries during the evaluation to finally output the best design from a set of candidate design particles.

\emph{Dimensional constrained} model-based optimization (CoMBO) considers a more general offline situation, where for a given initial design, optimizations are only allowed over a subspace of the whole space $\tilde{\mathcal{X}} \subset \mathcal{X}$.
Specifically, denote an initial design $\textbf{x} := \left[\textbf{x}^{opt} \| \textbf{x}^{con}\right]$ as the logical OR of two components, partial dimensions $\textbf{x}^{con}$ of $\textbf{x}$ are constrained to be fixed and treated as a constant $\textbf{c} \in \mathbb{R}^{\tilde{d}}$, only the other dimensions $\textbf{x}^{opt}$ are modifiable during optimization.
Therefore, the offline optimization is constrained in a subspace of design domain, $\tilde{\mathcal{X}}:= \{\textbf{x}\in \mathcal{X},\textbf{x} = \left[\textbf{x}^{opt}\|\textbf{x}^{con}\right]| \textbf{x}^{con}\in \mathbb{R}^{\tilde{d}}, x^{con}_i = c_i, i=1,2,\cdots,\tilde{d}, \tilde{d} < d\}$.
In the CoMBO setting, we study finding the best potential dimensions ${\textbf{x}^{opt}}^*$ for mediocre designs with dimensional constraints $\tilde{\mathcal{X}}$, which can be formulated as:
\begin{align}
    {\textbf{x}^{opt}}^* &\leftarrow \mathop{\arg\max}_{\textbf{x}^{opt}}f(\textbf{x}), \nonumber \\
    &\text{s.t. } \textbf{x} = \left[\textbf{x}^{opt}\| \textbf{c}\right], \textbf{x}\in\tilde{\mathcal{X}},
\end{align}
where $\|$ denotes the logical OR.
Without specific statements, we discuss the setting under an offline scenario, where optimization only allowed to make use of static offline dataset $\mathcal{D}=\{\textbf{x}_i,y_i\}^N_{i=1}$.

Compared to standard MBO, the CoMBO setting is more aligned with some real-world scenarios, making it valuable for study and discussion.
For example, in a combustion optimization process, the inlet air temperature, as a part of the external environment, is an important parameter for regulating boiler combustion efficiency, even though it cannot be altered;
in the problem of load balancing for mobile signal base stations, we typically adjust the cell individual offset (CIO) parameters rather than physical antenna tilt or transmission power, since making frequent changes to these physical parameters may be infeasible, etc.
In these real-world domains, black-box systems to be optimized often have some input dimensions that are directly constrained by the environment or are not desirable to be frequently modified.
Meanwhile, given any dimensional constraint, most of the designs in the offline dataset that satisfy that constraint are mediocre. Even though there exist points with high evaluation scores, these points often do not satisfy the given constraints. Points that have both high evaluation scores and satisfy the constraints are scarce in high-dimensional tasks, and in many cases, they are absent. Therefore, the idea of focusing on optimizing a mediocre initial design rather than attempting to find a solution closer to the global optimum is intuitive.
To this end, CoMBO is much better suited for addressing a significant portion of real-world scenario problems.\enlargethispage{20pt}

\subsection{Forward Surrogate Models}\label{sec:two-class-of-model}
To solve offline model-based optimization problem, a variety of MBO methods have been developed \cite{COMs, IOMs, CbAS, MINs, DDOM}, and the mainstream approaches can be broadly categorized into two types.

Inverse generative approaches learn a parameterized \emph{inverse} mapping from scores to conditional designs, $\hat{f}^{-1}_\theta(y, \textbf{z}) \rightarrow \textbf{x}$, where $y$ is a target score the model conditions on, and $\textbf{z}$ is usually a random signal \cite{CbAS, MINs, DDOM}. Such a one-to-many generative inverse model is usually non-derivable, and it is challenging to directly generate candidate designs that strictly satisfy dimensional constraints while leading to high evaluation scores in CoMBO setting.

In contrast, forward surrogate approaches learn a derivable \emph{forward} surrogate, which is the other indirect but effective option. To find the best possible solution, it trains a parametric model $\hat{f}_\theta(\textbf{x})$ as a forward surrogate of the black-box function $f(x)$, with an offline dataset $\mathcal{D}$ via supervised training:
$\hat{f}_\theta(\textbf{x}) \leftarrow \arg\min_\theta\sum_i(\hat{f}_\theta(\textbf{x}_i)-y_i)^2$ \cite{COMs, IOMs}.
Since the derivative information of such a parametric surrogate is accessible, it is easy to optimize $\textbf{x}$ against this learned surrogate. Typically, to find $\textbf{x}^*$, gradient descent starting from given initial design particles will be taken, as given by:
\begin{equation}
    \textbf{x}_{t+1} \leftarrow \textbf{x}_t + \eta\nabla_x \hat{f}_\theta(\textbf{x})|_{x=\textbf{x}_t}, \text{for } t\in\{1,\ldots, T\},
\end{equation}
where $\eta$ is the gradient step size, also known as the particle learning rate.
Such an approach naturally applies to the CoMBO setting, where the mediocre designs to be optimized with dimensional constraints are set as the initial design particles, and gradient steps over $\textbf{x}^{opt}$ are taken to derive ${\textbf{x}^{opt}}^*$.\enlargethispage{0pt}

\subsection{Retrieval-enhanced Machine Learning}

Retrieval techniques are widely adopted in machine learning scenarios \cite{zamani2022retrieval}, including language modeling \cite{zhao2022dense}, click-through rate prediction \cite{rim}, and disease risk assessment \cite{ma2022retrieval}. It has been verified that with the retrieved relevant data, the machine learning models commonly yield a better generalization performance. Generally, a $\theta$-parameterized machine learning model is denoted as $f_\theta(\mathbf{x})$; the corresponding retrieval-enhanced machine learning model is denoted as $f_\theta(\mathbf{x}, \mathcal{R}(\mathbf{x}))$,
where the target sample $\mathbf{x}$ is treated as a query for retrieval process,
$\mathcal{R}(\mathbf{x})$ is the retrieved relevant data instances.

\section{Methodology}
\begin{figure}[tbp]
    \centering
    \includegraphics[width=0.7\linewidth]{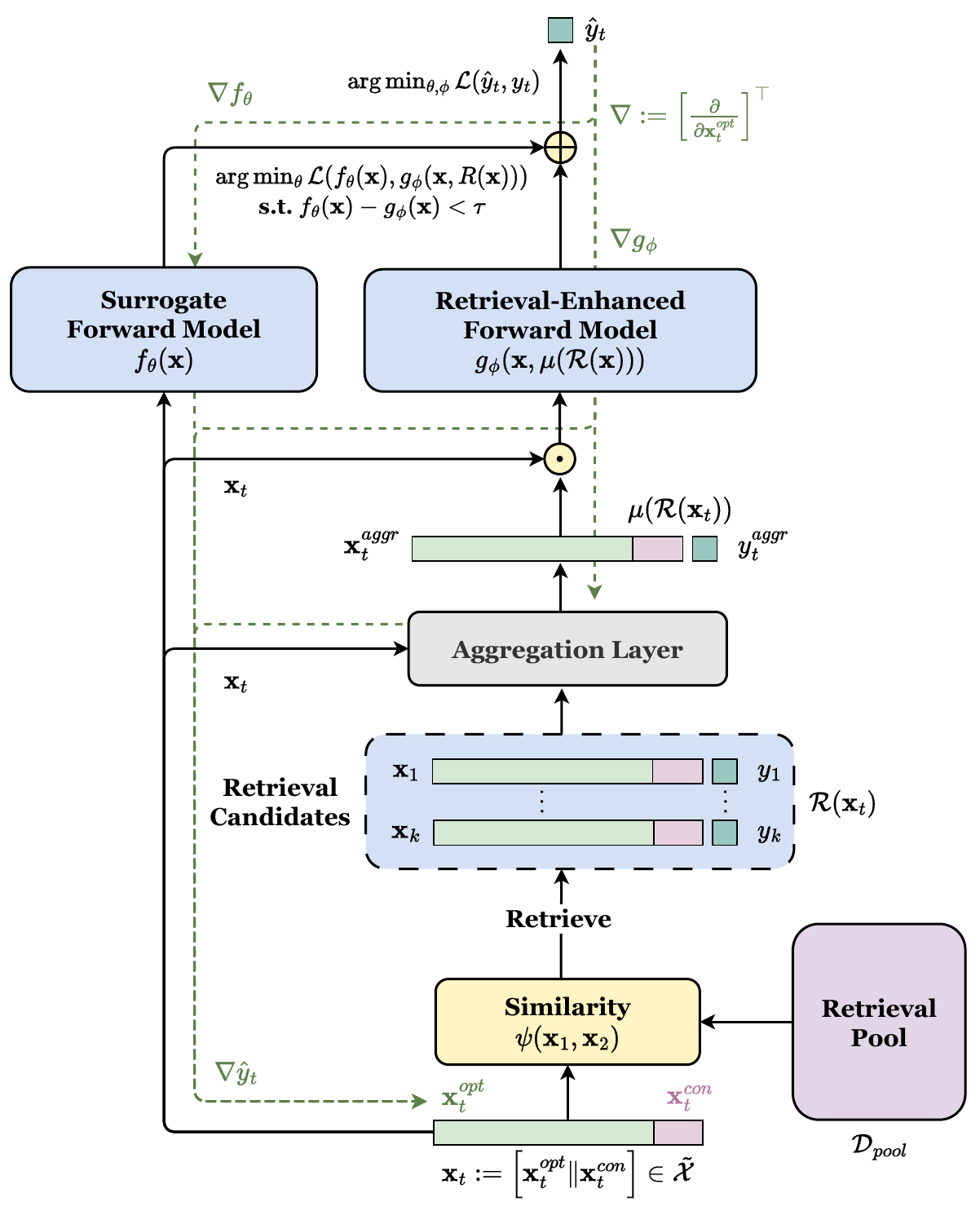}
    \caption{Overall framework of ROMO. The black solid lines indicate the forward inference process for an input design, the green dashed lines indicate the gradient propagation direction during gradient-based optimization, the ``$\odot$'' denotes concatenation, and the ``$\oplus$'' denotes weighted summation.}
    \label{fig:framework}
    \vspace{-15pt}
\end{figure}

In this section, we present our approach, \emph{retrieval-enhanced offline model-based optimization} (ROMO).
ROMO learns a retrieval-enhanced forward model that retrieves and aggregates in-support offline data to help estimate designs that could be out-of-distribution.
Based on the reference retrieval model, a surrogate forward model that provides stable gradient directions for optimization is built and regularized to produce estimates that keep consistent with the retrieval model while avoiding overestimation. As a result, when faced with designs from different score regions, ROMO can always make reliable estimates while also minimizing the impact of overestimation.
Figure~\ref{fig:framework} provides an overall framework.\pagebreak

\subsection{Over-conservatism in CoMBO}
Prior derivable surrogate approaches primarily concern the estimation accuracy and generalization capability of the model over inputs in the vicinity of high-score regions.
\citet{COMs, IOMs} have shown us that when starting from the top designs of the offline dataset, this class of approach can effectively identify candidates beyond the offline data support, and the truth scores of these found candidates can even be better than the best observation in the dataset.
These methods typically adopt additional pessimism and conservatism to ensure the model can safely generalize and extrapolate to OOD data, thus avoiding designs appearing erroneously good during the gradient-based optimization.\enlargethispage{18pt}

However, when it comes to the scenario of CoMBO, as we will show in Section~\ref{sec:hartmann}, these methods often fail since the potential over-conservatism can lead to new issues.

Specifically, for standard offline MBO, forward approaches only require setting top-$K$ offline points as initial design particles, and venturing further beyond the dataset support can always provide an opportunity to find promising candidate solutions.
Therefore, excessive conservatism typically does not lead to fatal issues but rather has a minor impact on the quality of the final candidate solutions, at most causing the omission of some better solutions.

When it comes to CoMBO, we attempt to set some mediocre designs as starting points and still try to search for the optimal solutions under the constraints of $\tilde{\mathcal{X}}$, leading to a very different situation:
Finding the optimal solution using gradient-based methods inevitably involves passing through a series of intermediate designs that are very likely to be OOD.
Excessive pessimism or conservatism will impede the gradient ascent process of design particles passing through OOD states, leading the algorithm to be trapped in local optima, even though the global optimal solutions may not necessarily be OOD.
Moreover, in such a new setting, we not only require the surrogate model to have good generalization capabilities around high-scoring region, but expect the surrogate model to provide highly reliable predictions for regions spanning from low to high scores because these intermediate points will evidently cover a broader region of design space.

To sum up, how to establish a high-fidelity forward surrogate across the entire space, while minimizing the occurrence of OOD issues, has become a critical challenge of CoMBO.

\subsection{Offline Data Retrieval}
The key idea behind our approach is to retrieve and use the relevant samples to assist the inference of the target sample.
In the field of information retrieval, neighbor-augmented deep models \cite{IR1, IR2, IR3, rim} show that integrating the feature and label representations of neighboring samples into the network can assist in predicting the target value.
Moreover, in the field of offline MBO, \citet{MINs} and \citet{DDOM} also confirm the effectiveness of data reweighting. By employing a simple reweighting strategy assigns higher training weights to data points located in high-score regions, these approaches can significantly enhance their generalization capabilities for high-score designs.
Inspired by these work, we propose to retrieve the offline dataset and dynamically construct a set of retrieval candidates for the current optimized design. We will show that such a set of retrieval candidates can significantly improve the optimization process.

In our approach, the entire offline dataset $\mathcal{D}=\{(\textbf{x}_i, y_i)\}^N_{i=1}, \textbf{x}\in\mathcal{X}, y\in \mathbb{R}$, is divided into three disjoint subsets as $\mathcal{D}_{train}, \mathcal{D}_{valid}$ and $\mathcal{D}_{pool}$, thus $\mathcal{D}_{train}\cup\mathcal{D}_{valid}\cup\mathcal{D}_{pool}=\mathcal{D}$.
The $\mathcal{D}_{train}$ and $\mathcal{D}_{valid}$ play a similar role to the training and validation sets in existing work.
The additional $\mathcal{D}_{pool}$, $|\mathcal{D}_{pool}|=M$, serves as a retrieval pool with size $M$, from which the retrieval enhanced model retrieves offline design-score pairs for $\textbf{x}_t$ and constructs a relevant set, $\mathcal{R}(\textbf{x}_t)=\{(\textbf{x}_i,y_i)\}^K_{i=1}$ with size $K$, of retrieval candidates.

Use a design feature $\textbf{x}_t$ as the query, to search relevant $(\textbf{x}, y)$ pairs from $\mathcal{D}_{pool}$, a similarity function $\psi(\cdot,\cdot)$ is necessary.
Function $\psi: (\mathcal{X},\mathcal{X}) \rightarrow \mathbb{R}$ defines the similarity metric between two samples, where samples with stronger correlations correspond to higher function values.
For the offline data retrieval process, it involves calculating the similarity between each sample in the retrieval pool $\mathcal{D}_{pool}$ and the query $\textbf{x}_t$  with a complexity of $\mathcal{O}(M)$.
Afterward, ranking the results with a complexity of $\mathcal{O}(M\log M)$ allows to filter out the most relevant top $K$ samples to construct a set of retrieval candidates $\mathcal{R}(\textbf{x}_t)$ for the current query $\textbf{x}_t$.

Therefore, the choice of $\psi$ is crucial, as it not only directly impacts the quality of retrieval candidates set but also determines the time complexity of offline data retrieval.
Depending on the task scenario, we have different choices for the metric function $\psi(\cdot,\cdot)$, such as inner product, kernel product, or micro-network \cite{product}.

For linearly separable tasks, the linear kernel's inner product is a simple but effective metric for distinguishing sample correlations, which is given by $\psi_{inner}(\textbf{x}_i,\textbf{x}_j) = \textbf{x}_i \cdot \textbf{x}_j$.
Meanwhile, the inner product offers a more efficient computation in high-dimensional tasks.
We find that using an RBF kernel product typically yields favorable results for most tasks,
despite the drawback of lower computational efficiency in higher dimensional tasks.
Cosine similarity yields good results as well in some low-dimensional tasks.

\subsection{Aggregation Learning}\label{sec:aggregation-learning}
A forward surrogate network typically has the form of $\hat{y} = \hat{f}_\theta(\textbf{x})$,
which is a function approximator of objective $f$ parameterized by $\theta$.
In this section, we discuss how to utilize the set of offline retrieval candidates $\mathcal{R}(\textbf{\textbf{x}})$ to assist the score prediction of forward model $\hat{f}_\theta$.

According to this goal, an aggregation learning process, which we denote by $\mu(\cdot)$, is introduced to aggregate information from $\mathcal{R}(\textbf{x})$ and obtain the aggregated retrieved representation $\mu(\mathcal{R}(\textbf{x}))$. The calculation of representation is defined by
\begin{align}
    \mu(\mathcal{R}(\textbf{x}_t)) &= (\textbf{x}^{aggr}_t, y^{aggr}_t), \nonumber \\
    \textbf{x}^{aggr}_t &= \sum\nolimits^{K}_{i=1}w_i\cdot \textbf{x}_i, \nonumber \\
    y^{aggr}_t &= \sum\nolimits^{K}_{i=1}w_i\cdot y_i, (\textbf{x}_i, y_i)\in\mathcal{R}(\textbf{x}_t),
\end{align}
where the derived representation contains feature $\textbf{x}^{aggr}$ and score $y^{aggr}$ part. The aggregation process can be considered as a weighted summation over samples in the set of retrieval candidates, where the $\textbf{x}$ and $y$ share the same aggregation weights $\{w_i\}^K_{i=1}$.

The process of obtaining this set of weights in ROMO can be either parametric or non-parametric.
For the parametric method, we learn this set of weights using an attention-style approach.
In \citet{rim}, the aggregation weight $w_i$ is learned by an attention layer parameterized by parameter matrix $\textbf{A}$.
The dot product between query $\textbf{A}^\top_q\textbf{x}_t$ and keys $\{\textbf{A}^\top_k\textbf{x}_i\}^K_{i=1}$ is computed and mapped to attention weights utilizing the softmax function.

We notice that, for such a dot-product attention layer, the usage of softmax makes the learned aggregation weights always positive and sum up to 1. Therefore, the aggregated representation always lies within the convex hull of the retrieval set.

\begin{figure}[tbp]
    \centering
    \includegraphics[width=0.9\linewidth]{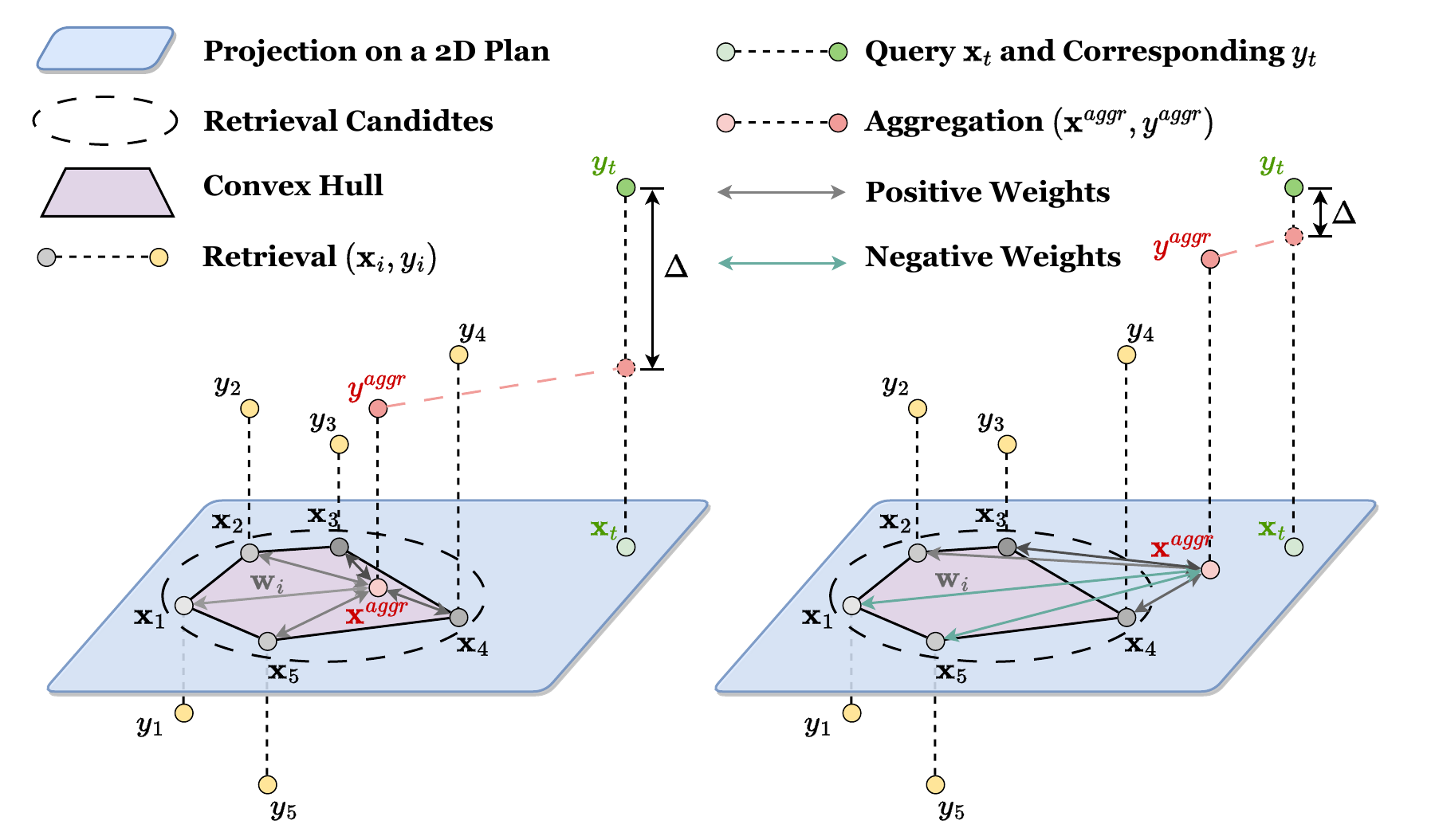}
    \caption{An illustration of how the negative aggregation weights can help. Left: The diagram shows that when the $\textbf{w}_i$ is always positive, the $\textbf{x}^{aggr}$ must fall within the convex hull of a set of retrieval candidates, which may lead to undesirable outcomes $y^{aggr}$. Right: The diagram shows the case where allowing $\textbf{w}_i$ to be negative, and we can obtain an affine $\textbf{x}^{aggr}$ outside the convex hull, thus achieving more reliable $y^{aggr}$.}
    \label{fig:aggregation}
    \vspace{-15pt}
\end{figure}

This characteristic can lead to some problems.
For an offline CoMBO approach like ROMO, the current query $\textbf{x}_t$ may potentially be an OOD point, meaning that $\textbf{x}_t$ as an affine combination of retrieved samples falls outside the convex hull of the set of retrieval candidates.
As shown in Figure~\ref{fig:aggregation}, in this scenario, the aggregation weights obtained using the softmax function, applied to aggregate the representation $\textbf{x}^{aggr}$, may exhibit significant differences from the query $\textbf{x}_t$.
Consequently, the corresponding aggregation label $y^{aggr}$ cannot guarantee reliable guidance for predicting $y_t$.

To mitigate the aforementioned issue, we propose to allow aggregation weights to be negative under the constraint of $\|\textbf{w}\|_1=1$, and enable a retrieval-enhanced model dealing with the OOD query.
One mild way to achieve this goal is to apply an appropriate transformation to the softmax function.
We apply a slight scaling to the range of softmax values, extending the coverage of function values, and then slightly shifting it towards the negative axis without altering the expectations and L1-norms of the weights as
\begin{equation}\label{eq:gamma}
    \text{Softmax}'(\textbf{x}_K^\top\textbf{A}\textbf{x}_t) := \gamma \text{Softmax}\left(\frac{\textbf{x}_K^\top\textbf{A}\textbf{x}_t}{\sqrt{d}}\right) - \frac{\gamma - 1}{K},
\end{equation}
where $\textbf{x}_K^\top\textbf{A}\textbf{x}_t$ is the attention scores, $\textbf{x}_K\in\mathbb{R}^{d\times K}$ is the retrieval candidates with size $K$, $\textbf{x}_t \in \mathbb{R}^d$ is the query, and $\gamma$ is a tuning parameter.
The range of the $\text{Softmax}'(\cdot)$ thus becomes $[-\frac{\gamma-1}{K},\gamma-\frac{\gamma-1}{K}]$ with an expectation of $\frac{1}{K}$.
This type of aggregation learning approach includes a learnable attention layer. Therefore, we refer to it as a parametric aggregation learning implementation.

For non-parametric methods, a more direct implementation approach is to explicitly impose a hard constraint of $\|\textbf{w}\|_1=1$ on linear equation $\textbf{x}_K\textbf{w} = \textbf{x}_t$ and directly derive the closed-form solution for $\textbf{w}$.
Specifically, to determine the values of $\textbf{w}$, we employ a ridge regression-like solver as
\begin{equation}
    \hat{\textbf{w}}_\lambda \leftarrow \mathop{\arg\min}_\textbf{w} \|\textbf{x}_t - \textbf{x}_K\textbf{w}\|^2_2 + \lambda\|\textbf{w}\|^2_2, \label{eq:solve-ridge}
\end{equation}
where the $\lambda > 0$ is a ridge parameter, and regularization term $\lambda\|\textbf{w}\|_2^2$ is introduced to provide a robust solution.
Notice that, the closed form solution $\hat{\textbf{w}}_\lambda = (\textbf{x}_K^\top\textbf{x}_K + \lambda \textbf{I}_p)^{-1}\textbf{x}_K^\top\textbf{x}_t$ given by Eq.~\eqref{eq:solve-ridge}
is an approximate solution to $\textbf{x}_K\textbf{w}=\textbf{x}_t$ and it cannot guarantee $\|\textbf{w}\|_1=1$. 

Considering that aggregation $\textbf{x}^{aggr}$ does not need to be identical to $\textbf{x}_t$ since $\textbf{x}_t$ can be directly concatenated into the input of the retrieval-enhanced model.
In contrast, aggregation $y^{aggr}$ will provide crucial assistance for the predictions of the retrieval-enhanced model,
We normalize $\hat{\textbf{w}}_\lambda$ to obtain the final $\textbf{w}$ as
\begin{equation}
    \textbf{w} = \frac{(\textbf{x}_K^\top\textbf{x}_K + \lambda \textbf{I}_p)^{-1}\textbf{x}_K^\top\textbf{x}_t}{\|(\textbf{x}_K^\top\textbf{x}_K + \lambda \textbf{I}_p)^{-1}\textbf{x}_K^\top\textbf{x}_t\|_1},
\end{equation}
thus the $w_i$'s sum up to 1.

This approach of solving a linear equation and utilizing the closed-form solution as the aggregation weights is a deterministic process without learnable parameters. We refer to it as a non-parametric aggregation learning implementation.

With the learned aggregation weights $\textbf{w}$, the aggregated retrieved representation $\mu(\mathcal{R}(\textbf{x}))$ can be derived.
We can augment the input of forward model $\hat{f}_\theta(x)$ by concatenate the $\mu(\mathcal{R}(\textbf{x}))$ with original input $\textbf{x}$, and build a retrieval-enhanced forward model as $\hat{y} = \hat{f}_\theta(\textbf{x},\mu(\mathcal{R}(\textbf{x})))$.

Hence, utilizing offline training data $\mathcal{D}_{train}$, the parameter $\theta$ of the retrieval-enhanced forward model can be  learned by supervised regression as
\begin{equation}
    \hat{f}_\theta(\textbf{x},\mu(\mathcal{R}(\textbf{x}))) \leftarrow \mathop{\arg\min}_\theta \sum\nolimits_{\textbf{x} \in\mathcal{D}_{train}}\mathcal{L}_{mse}(\hat{f}_\theta(\textbf{x},\mu(\mathcal{R}(\textbf{x}))), y).
\end{equation}

\subsection{Retrieval-enhanced Surrogate Optimization}
The use of aggregations enables the retrieval-enhanced forward model to provide more reliable predictions for the query $\textbf{x}_t$, even when $\textbf{x}_t$ is out-of-distribution. 
In such cases, the presence of $y^{aggr}$ offers guidance for predicting $\textbf{x}_t$, preventing the output score from becoming erroneously too high or too low.

However, directly applying gradient-based optimization on the retrieval-enhanced forward network $f(\textbf{x},\mu(\mathcal{R}(\textbf{x})))$ to optimize mediocre designs using the partial derivative of $\hat{y}$ w.r.t. $\textbf{x}^{opt}$ is usually not feasible.
Due to the inclusion of $y^{aggr}$, $\mu(\mathcal{R}(\textbf{x}))$ often exerts strong guidance on the retrieval model's predictions while also dominating the gradient of $f(\textbf{x},\mu(\mathcal{R}(\textbf{x})))$.
As a result, when the gradient ascent is applied, the gradient components are primarily concentrated on $\mu$ while being particularly tiny on $\textbf{x}$.
In CoMBO, when we only want to optimize the dimensions in $\textbf{x}$ that belong to $\textbf{x}^{opt}$ and mask out the gradients for the parts of $\textbf{x}^{con}$, this problem can become even more severe.

To address this issue, in addition to training a retrieval-enhanced forward model, we employ another forward surrogate model.
We denote the surrogate network as $\hat{f}_\theta(\textbf{x})$, parameterized by parameters $\theta$, and denote the retrival-enhanced network as $\hat{g}_\phi(\textbf{x},\mu(\mathcal{R}(\textbf{x})))$,  parameterized by parameters $\phi$.
$f_\theta$ only accepts the query design $\textbf{x}$ as input and outputs the predicted score; therefore the gradient of $f_\theta$ w.r.t. $\textbf{x}^{opt}$ will not be influenced by the aggregations in the surrogate forward model, 
whereas $g_\phi$ accepts additional aggregated retrieved information and provide a reference prediction for $f_\theta$.

In practical implementation, we combine the predictions from these two networks by weighted summation to obtain the final prediction of ROMO: $\hat{h}_{\theta,\phi}(\textbf{x}_t) = \hat{y}_t = \beta f_\theta(\textbf{x}_t) + (1 -\beta)g_\phi(\textbf{x}_t, \mu(\mathcal{R}(\textbf{x}_t)))$,
where the $\beta \in \left(0, 1\right)$ is a parameter that weighted predictions from $f_\theta$ and $g_\phi$.
To simplify the training process, we set $\beta = 0.5$ and jointly train the networks through supervised learning:
\begin{align}\label{eq:train}
    \hat{h}^*_{\theta,\phi} \leftarrow \mathop{\arg\min}_{\theta\in\Theta, \phi \in \Phi} \frac{1}{2}\overbrace{\mathbb{E}_{(\textbf{x},y)\sim\mathcal{D}_{train}}\left[\left(\hat{h}_{\theta,\phi}(\textbf{x})-y\right)^2\right]}^{\mathcal{L}_{s}:\text{~standard supervised regression}} \nonumber \\
    + \overbrace{\alpha \left(\mathbb{E}_{\textbf{x}\sim\mathcal{D}_{train}}\left[\hat{f}_\theta(\textbf{x})\right] - \mathbb{E}_{\textbf{x}\sim\mathcal{D}_{train}}\left[\hat{g}_\phi(\textbf{x},\mu(\mathcal{R}(\textbf{x})))\right]\right)}^{\mathcal{L}_{c}:\text{~conservatism regularizer}} \nonumber \\
    + \underbrace{\mathbb{E}_{\textbf{x}\sim \mathcal{D}_{train}}\left[\left(\hat{f}_\theta(\textbf{x})-\hat{g}_\phi(\textbf{x},\mu(\mathcal{R}(\textbf{x})))\right)^2\right]}_{\mathcal{L}_{a}:\text{~alignment regularizer}}.
\end{align}

For the training objective in Equation~\eqref{eq:train}, the standard supervised regression term $\mathcal{L}_s$ leads to a regression process by the least squares method. The alignment regularizer $\mathcal{L}_{a}$ is designed to constrain the outputs of $\hat{f}_\theta$ aligning with the reference predictions made by $\hat{g}_\phi$. Without the alignment regularizer term, the $f_\theta$ is prone to learn the residual between $\hat{g}_\phi(\textbf{x}_t)$ and corresponding $y_t$, rather the value of $y_t$, since the prediction of retrieval-enhanced forward model $\hat{g}_\phi$ often makes a more accurate estimate of $y$ with the guidance from aggregation. The conservatism regularizer $\mathcal{L}_c$ is introduced to encourage the surrogate forward model $\hat{f}_\theta$ makes a conservative prediction that no greater than reference value given by retrieval-enhanced forward model $\hat{g}_\phi$. We find the explicit conservatism regularizer is much more effective than a quantile loss in alignment regularizer term that encourages conservatism implicitly.

Selecting a single value of $\alpha$ that works for most tasks is difficult. Considering that, we introduce a training trick that is used in \citet{COMs} to modify the training objective as
\begin{align}
    \hat{h}^*_{\theta,\phi} \leftarrow \mathop{\arg\min}_{\theta\in\Theta, \phi \in \Phi} \frac{1}{2}\mathbb{E}_{(\textbf{x},y)\sim\mathcal{D}_{train}}\left[\left(\hat{h}_{\theta,\phi}(\textbf{x})-y\right)^2\right] \quad\quad\nonumber \\
    + \mathbb{E}_{\textbf{x}\sim \mathcal{D}_{train}}\left[\left(\hat{f}_\theta(\textbf{x})-\hat{g}_\phi(\textbf{x},\mu(\mathcal{R}(\textbf{x})))\right)^2\right], \quad\quad \label{eq:train-dual} \\
    \text{s.t.~} \left(\mathbb{E}_{\textbf{x}\sim\mathcal{D}_{train}}\left[\hat{f}_\theta(\textbf{x})\right] - \mathbb{E}_{\textbf{x}\sim\mathcal{D}_{train}}\left[\hat{g}_\phi(\textbf{x},\mu(\mathcal{R}(\textbf{x})))\right]\right) \leq \tau .\nonumber
\end{align}
In Eq.~\eqref{eq:train-dual}, a less task-sensitive parameter, $\tau$, is involved in the constraint, replacing the parameter $\alpha$ in Eq.~\eqref{eq:train}. We can solve the optimal $(\theta,\phi)$ under the constraint given by $\tau$ with a process of dual gradient descent for this new training objective.

\subsection{Insights about ROMO Behavior}
To better understand why ROMO can generalize better for input either from the low-scoring or the high-scoring region, we provide some insights into the behavior of retrieval-enhanced models.

Consider a retrieval-enhanced forward model with 2-layer networks as
$g(\mathbf{x}_{in}; \phi) = \sum_{k=1}^{K} a_k \sigma (\mathbf{b}^\top_k\mathbf{x}_{in}+c_k),$
where input $\mathbf{x}_{in}$ contains original query $\mathbf{x}^o$, aggregation feature $\mathbf{x}^{aggr}$, and aggregation label $\mathbf{y}^{aggr}$, $K$ is the number of hidden nodes, $\sigma(\cdot)$ denotes the activation function, so that $\mathbf{\phi} = (\mathbf{a},\mathbf{c},\mathbf{b}_1, \ldots, \mathbf{b}_k)$ denotes all the parameters.
We decompose the $\mathbf{x}_{in}$ into two partitions of $\mathbf{x} := \left[\mathbf{x}^{o}\|\mathbf{x}^{aggr}\right]$ and $y$, which correspond to parameter $\phi^{\mathbf{x}}$ and $\phi^{\mathbf{y}}$, $\phi = \phi^{\mathbf{x}} + \phi^y$. Thus, the retrieval-enhanced 2-layer network can be rewritten as
\begin{align}
    & g(\mathbf{x}^o,\mathbf{x}^{aggr},y^{aggr};\mathbf{\phi}^{\mathbf{x}},\phi^{y}) \nonumber \\
    :=& \sum\nolimits_{i=1}^K a^\mathbf{x}_k \sigma({\mathbf{b}^\mathbf{x}_k}^\top \textbf{x} + c_k^\mathbf{x}) + \sum\nolimits_{i=1}^K a_k^y \sigma(\mathbf{b}_k^y y^{aggr} + c_k^y) \\
    =& g(\mathbf{x}^o, \mathbf{x}^{aggr};\phi^\mathbf{x}) + g(y^{aggr};\phi^y). \nonumber 
\end{align}
Assume the least square loss is used to fit this network $f$ on offline data $\{\mathbf{x}_i,y_i\}^N_{i=1}$, the process can be represented by minimizing
\begin{align}
    \mathcal{R}_{emp}(\phi^\mathbf{x},\phi^y) :=& \frac{1}{N}\sum\nolimits^N_{i=1}(g(\mathbf{x}_i^o,\mathbf{x}_i^{aggr},y_i^{aggr};\mathbf{\phi}^{\mathbf{x}},\phi^{y}) - y_i)^2 \\
    = \frac{1}{N}&\sum\nolimits^N_{i=1}\left[g(\mathbf{x}_i^o,\mathbf{x}_i^{aggr};\mathbf{\phi}^{\mathbf{x}}) - \left(y_i - g(y^{aggr}; \phi^y)\right)\right]^2.\nonumber 
\end{align}

Note that the aggregation label $y^{aggr}$ obtained using the aggregation learning approach proposed in Section~\ref{sec:aggregation-learning} is typically considered to exhibit high consistency with the values held by $y_i$.
This makes $\phi^y$ yield a more consistent update direction compared to $\phi^\mathbf{x}$ when using a gradient-based optimizer, like SGD, to update the parameters of the network $g$.
As a result, $\phi^y$ can converge more quickly than $\phi^\mathbf{x}$, and tend to learn to output a value $g(y^{aggr};\phi^y)=\hat{y}_i$ that lies very close to label $y_i$ after a few steps of training.
We formulate it as $y_{\delta i} := |\hat{y}_i - y_i| < \delta$, a residual between $y_i$ and $\hat{y}_i$, where $\delta$ is a small value.
Therefore, shortly after the training begins, fitting $\{\mathbf{x}_i, y_i\}_{i=1}^N$ becomes minimizing $\mathcal{R}_{emp}(\phi^\mathbf{x}) := \frac{1}{N}\sum^N_{i=1}(g(\mathbf{x}_i^o,\mathbf{x}_i^{aggr};\mathbf{\phi}^{\mathbf{x}}) - y_{\delta i})^2$, which can be regarded as learning $\phi^\mathbf{x}$ to fit a tiny residual of $y_{\delta i}$.

With the help of common deep learning tricks such as near-zero initialization, early stopping, learning rate decay, weight decay, etc., in order to fit the tiny residual, the parameter optimizer of the retrieval-enhanced model will keep exploring the area in a small vicinity of zero \cite{generalize1, generalize2, generalize3}.
According to Theorem~1 as provided in \citet{generalize}, if we utilize the term $\mathbb{E}\|\nabla_\mathbf{x}g(\mathbf{x})\|^2_2$ to measure the complexity of learned retrieval-enhanced model $g(\mathbf{x}^o,\mathbf{x}^{aggr};\phi^\mathbf{x})$ by
\begin{equation}
    \mathbb{E}\|\nabla_\mathbf{x}g(\mathbf{x})\|^2_2 = \sum\nolimits_{k1, k2} \mathbf{b}^\top_{k_1} \mathbf{b}_{k_2} I_{\mathbf{c}}(k_1, k_2),
\end{equation}
where $I_\mathbf{c}$ is the Fisher information matrix w.r.t. model parameters $\mathbf{c}$,
then the complexity metric of the network will be bounded by
\begin{equation}
    2 \mathbb{E}\|\nabla_\mathbf{x}g(\mathbf{x})\|^2_2 \leq \|B\|^4_F + \|I_\mathbf{c}\|^2_F,
\end{equation}
where $B = (\mathbf{b}_1,\ldots,\mathbf{b}_K)\in \mathbb{R}^{d\times K}$.
That is, in a small near-zero parameter space given by $\mathcal{C} := \{\phi | \|B\|^4_F + 2\max_k |a_k| \|\mathbf{b}_k\|^2_2 \leq \eta\}$, with $\eta$ being very small, a low-complexity solution with small Hessian could always be found.

Therefore, fitting the residual and exploring the area in the small vicinity of zero help find a low-complexity solution in the flat and large attractor basins.
Within such an area, the high-complexity minima generalizing, like random guessing, has much smaller attractor basins; empirically, we never observe that optimizers converge to these bad solutions, even though they exist in parameter space \cite{generalize4}. 
As such, we can explain why the generalization ability of a retrieval-enhanced model is better.

Benefiting from the retrieval-enhanced network's strong generalization, we can utilize it to provide a reliable reference estimate.
The surrogate forward model learns to make predictions bounded by retrieval-enhanced estimates, further reducing the risk of overestimation,
and providing effective gradients for optimization.

\section{Experiments}\label{sec:experiments}
In this section, we first provide the offline evaluation protocol we use for the CoMBO setting, and then present three different types of experimental settings and corresponding results in detail\footnote{Our implementation of ROMO can be found at \url{https://github.com/cmciris/ROMO}.}.

\subsection{Evaluation for CoMBO}
For a CoMBO task, we define the mediocre samples set as $\tilde{\mathbf{x}} = \{\mathbf{x} | (\mathbf{x}, y)\in\mathcal{D}, y<y_{thres}\}$, where samples with scores less than a threshold $y_{thres}$ will be considered as a mediocre one.
Typically, we can use bottom $K$ samples found by $\arg\min_{(\mathbf{x},y)\in\mathcal{D}} y$ to represent $\tilde{\mathbf{x}}$ as well.
CoMBO aims to optimize such a mediocre sample set with only $\textbf{x}^{opt}$ as modifiable. When using gradient-based optimization, $\tilde{\mathbf{x}}$ serves as the initial design particles, and $T$ steps of gradient ascent are applied to $\tilde{\mathbf{x}}^{opt}$ to generate potential optimal solutions.
We propose the following two protocols to determine the values of $T$ and how to select candidates from a series of derived solutions.

\minisection{Protocol 1}
In an ideal scenario, we train the surrogate to fit the training dataset completely. Then, use a sufficiently large $T$ to perform optimization with the deriving surrogate until the solution converges.
Solutions with the best predicted scores are selected and evaluated via querying the black-box function.

\minisection{Protocol 2}
Since predicted scores often exhibit significant fluctuations in a complex task in real-world scenarios, it is challenging to precisely locate solutions with the best-predicted score as in ideal scenarios.
We use a limited $T$ and exhaust all $T$ steps of gradient-based optimization.
Similar to the standard offline MBO setting, in CoMBO, we also allow for an evaluation budget of size $Q$. All the solutions produced in the last $Q$ optimization steps are evaluated via the truth function, and the best score is reported.

\subsection{Compared Methods}
We compare ROMO with the mainstream forward surrogate approaches.
\textbf{Gradient Ascent} - a naive fully connected forward surrogate;
\textbf{IOM} - an invariant objective model proposed in \citet{IOMs};
\textbf{COMs} - conservative objective models proposed in \citet{COMs};
\textbf{REM}$_p$ - a retrieval-enhanced model only, with a trainable attention layer;
\textbf{REM}$_n$ - a retrieval-enhanced model only, with a ridge aggregation learning;
\textbf{ROMO}$_p$ - our proposed approach, with a retrieval-enhanced model REM$_p$;
\textbf{ROMO}$_n$ - our proposed approach, with a retrieval-enhanced model REM$_n$.

\subsection{Hartmann (3D) Test Function}\label{sec:hartmann}
The Hartmann function refers to a family of functions with multiple local minima. It is commonly used as a benchmark in optimization problems, and finding the global minimum is typically challenging.

To evaluate the performance of ROMO under an offline CoMBO setting, we employ a Hartmann (3D) function with four local maxima and one global maximum, and evaluate it on the hypercube $\mathcal{X} = \{\mathbf{x}|x_i\in (0, 1), i=0,1,2\}$.
We uniformly sample 12k points in $\mathcal{X}$, remove the top and bottom 1k points with the highest and lowest function values, and use the remaining 10k points to construct the offline dataset $\mathcal{D}$.
To construct a dimensional constraint $\tilde{\mathcal{X}}$ in the task, given a point to be optimized, we fix $x_2$ and allow modifications only in the $x_0$ and $x_1$ dimensions.

\begin{table}[tbp]
    \centering
    \caption{Unnormalized score for Hartmann.}
    \vspace{-8pt}
        \resizebox{0.72\linewidth}{!}{
        \begin{tabular}{llll}
            \toprule
             & \textbf{Mean} & \textbf{Maximum} & \textbf{Median} \\
            \midrule
            Grad. & 0.995±0.094 & 2.592±0.199 & 0.875±0.137 \\
            IOM & 0.749±0.202 & 2.802±0.310 & 0.565±0.267 \\
            COMs & 1.127±0.069 & 2.683±0.188 & 1.033±0.065 \\
            REM$_p$ & 0.738±0.486 & 2.244±1.282 & 0.547±0.559 \\
            REM$_n$ & 1.135±0.136 & 3.590±0.092 & 0.522±0.254 \\
            \midrule
            \textbf{ROMO}$_p$ & \underline{1.758±0.117} & \underline{3.600±0.325} & \underline{2.148±0.265} \\
            \textbf{ROMO}$_n$ & \textbf{1.823}±\textbf{0.086} & \textbf{3.721}±\textbf{0.136} & \textbf{2.294}±\textbf{0.171} \\
            \midrule
            $\tilde{\mathbf{x}}$ & 0.120 & 0.281 & 0.108 \\
            \bottomrule
        \end{tabular}}
    \label{tab:hartmann}
\end{table}

\minisection{Overall Performance}
We divide the offline dataset into 50 bins uniformly based on the values of $x_2$, select the bottom-2 samples from each bin to form the initial design particles $\tilde{\mathbf{x}}$, and then use the protocol 1 to perform the evaluation. The overall performance is shown in Table~\ref{tab:hartmann}, where the mean value, best value, and median value of $\tilde{\mathbf{x}}_T$ are reported.
In the Hartmann task, whether using a learnable attention layer or a ridge aggregation, ROMO exhibits the best mean, maximum, and median performance. COMs has the second-best performance, but there is still a significant gap compared to ROMO.
Although the remaining methods can achieve competitive maximum performance, they perform poorly in terms of mean and median values.
It is worth noting that, despite using the same retrieval-enhanced network, REM does not perform well.

\aptLtoX[graphic=no,type=html]{
\begin{figure*}[htbp]
    \centering
    \subfloat[Gradient Ascent]{
        \centering
        \includegraphics[width=0.4\columnwidth]{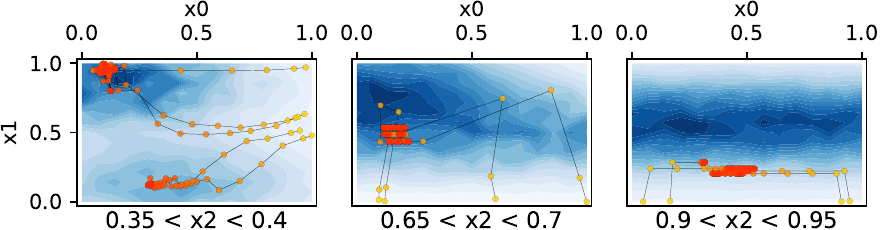}
    }
    \subfloat[COMs]{
        \centering
        \includegraphics[width=0.4\columnwidth]{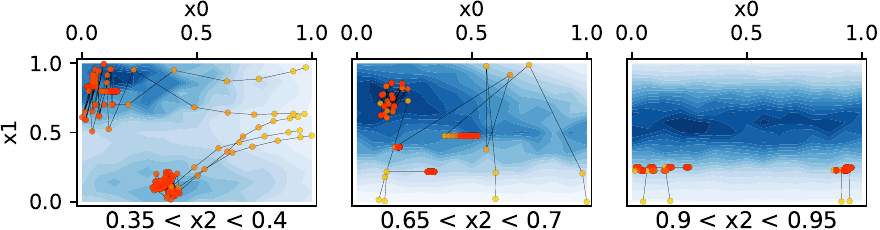}
    } \\
    \vspace{-10pt}
    \subfloat[REM]{
        \centering
        \includegraphics[width=0.4\columnwidth]{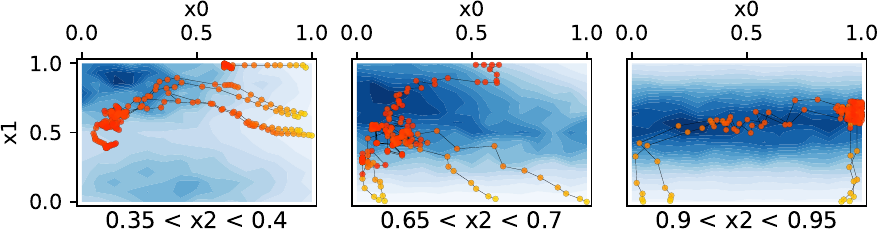}
    }
    \subfloat[ROMO]{
        \centering
        \includegraphics[width=0.4\columnwidth]{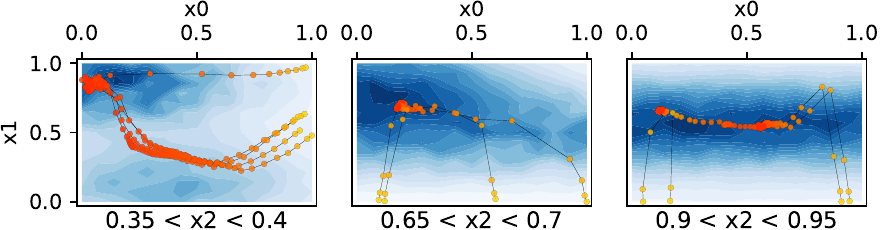}
    }
    \vspace{-10pt}
    \caption{Visualization of different model behaviors. The background shows Hartmann's ground truth manifold of different bins. The points, transitioning from yellow to red, connected by black solid lines, record the positions of solutions produced during the $T=50$ steps gradient ascent process, with yellow representing the starting point and red representing the endpoint. }
    \label{fig:model-behavior}
\end{figure*}}{
\begin{figure*}[htbp]
    \centering
    \subfloat[Gradient Ascent]{
        \centering
        \includegraphics[width=0.9\columnwidth]{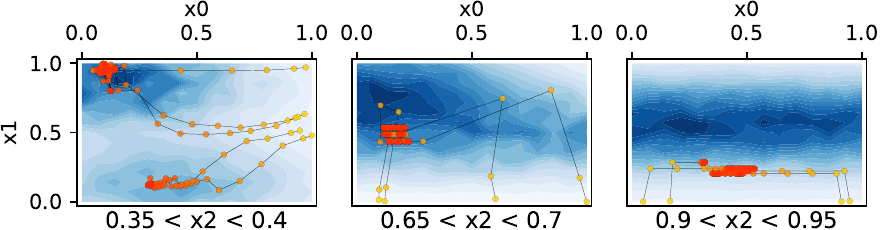}
    }
    \subfloat[COMs]{
        \centering
        \includegraphics[width=0.9\columnwidth]{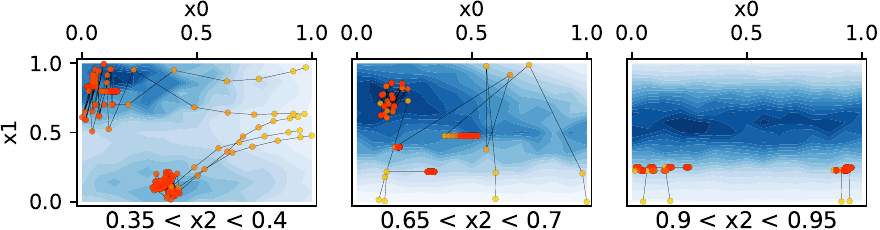}
    } \\
    \vspace{-10pt}
    \subfloat[REM]{
        \centering
        \includegraphics[width=0.9\columnwidth]{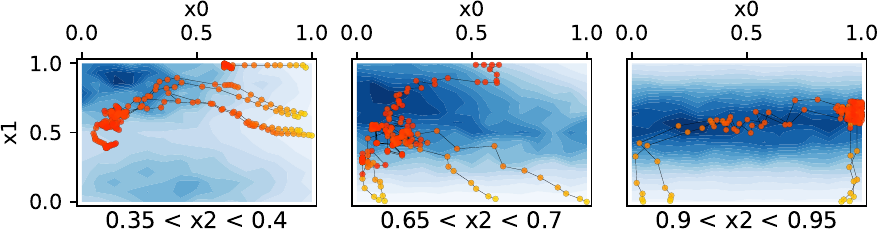}
    }
    \subfloat[ROMO]{
        \centering
        \includegraphics[width=0.9\columnwidth]{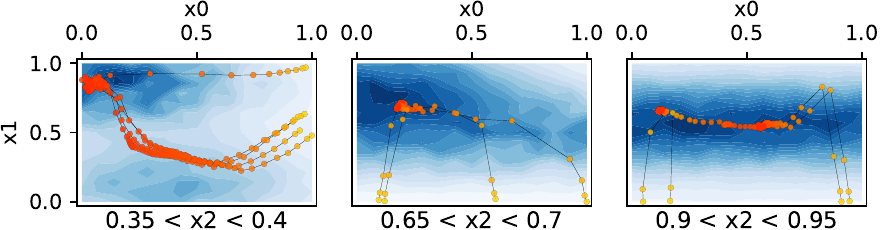}
    }
    \vspace{-10pt}
    \caption{Visualization of different model behaviors. The background shows Hartmann's ground truth manifold of different bins. The points, transitioning from yellow to red, connected by black solid lines, record the positions of solutions produced during the $T=50$ steps gradient ascent process, with yellow representing the starting point and red representing the endpoint. }
    \label{fig:model-behavior}
\end{figure*}}

\minisection{Model Behavior}
To understand what causes differences in optimization performance, we analyze the behavior of different models throughout the optimization process from starting point $\tilde{\mathbf{x}}$ to the final result $\tilde{\mathbf{x}}_T$.
As shown in Figure~\ref{fig:model-behavior}, Gradient Ascent and ROMO both exhibit favorable behavior in approaching global or local optima points.
The manifold learned by Gradient Ascent seems to be sharper, which leads the optimization of Gradient Ascent to be prone to converge at the nearest suboptimal local maxima.
COMs often get stuck in some non-local optimal positions. We speculate that this could be due to the overly strong constraints introduced by COMs, leading to over-underestimation of certain points on the manifold. These over-underestimations create erroneous local maxima, trapping data points from further optimizing in the correct direction.
ROMO performs well in most cases, and sometimes even shows the ability of detouring to achieve the global optima.
Surprisingly, despite the difference in whether an additional forward surrogate model is used or not, the behavior of REM is far from satisfactory. We find that although in some cases the points converge to the correct position, REM often exhibits gradient steps that drift uncontrollably toward an undesirable direction. This is due to the usage of additional dimensions $\mu(\mathcal{R}(\mathbf{x}))$, which not only dilutes the gradient components in the $\mathbf{x}^{opt}$ direction but may also lead to a problem where the gradient is dominated by $y^{aggr}$.

\subsection{Cell Individual Offset Tuning Task}
Cell individual offset (CIO) tuning task is a real-world CoMBO task, where the physical parameters like antenna tilt and transmission power of 4G mobile signal base stations are fixed, and only the CIO parameters are adjustable to balance the load of base stations.
In such a load-balancing task, the optimization objective is set as the negative variance of the load among the base stations. Load imbalances are thus considered mediocre points, and we aim to optimize these points to maximize the objective.

The CIO dataset used in this paper contains 100k samples. Each sample consists of 138 dimensions of continuous features, where only 36 dimensions of CIO parameters are adjustable. We remove the top and bottom 1k samples, and then evaluate over the remaining samples. Protocol 2 is used, with $T=250$, $Q=10$, and $|\tilde{\mathbf{x}}|=256$.

\begin{table}[tbp]
    \centering
    \caption{Normalized score for CIO Tuning.}
    \vspace{-8pt}
        \resizebox{0.78\linewidth}{!}{
        \begin{tabular}{llll}
            \toprule
             & \textbf{Mean} & \textbf{Maximum} & \textbf{Median} \\
            \midrule
            Grad. & 93.151±1.763 & \textbf{101.019}±\textbf{1.667} & 93.459±1.870 \\
            IOM & 92.021±1.410 & 98.287±1.510 & 92.052±1.284 \\
            COMs & 92.199±0.434 & 98.642±1.234 & 92.265±0.456 \\
            REM$_p$ & 92.760±1.711 & 99.852±1.288 & 93.246±1.684 \\
            REM$_n$ & 93.079±1.660 & \underline{99.996±1.097} & 93.399±1.603 \\
            \midrule
            \textbf{ROMO}$_p$ & \underline{93.593±0.684} & 99.084±0.948 & \underline{93.868±0.711} \\
            \textbf{ROMO}$_n$ & \textbf{94.425}±\textbf{1.087} & 99.415±0.933 & \textbf{94.586}±\textbf{0.996} \\
            \midrule
            $\tilde{\mathbf{x}}$ & 82.016 & 83.474 & 82.010 \\
            \bottomrule
        \end{tabular}}
    \label{tab:cio}
\end{table}

We report the normalized scores instead of unnormalized ones since the objective value of negative variance is usually small.
Table~\ref{tab:cio} demonstrates that,
all compared methods achieve competitive performance with sufficient offline data and non high-dimensional features. ROMO demonstrates the best mean and median performance, while Gradient Ascent discover better solutions.

\subsection{Modified Design-Bench Tasks}
Although the standard offline MBO benchmark, Design-Bench \cite{design-bench}, is not designed for CoMBO, we provide experiments on Hopper Controller and TF Bind 8, one continuous and one discrete task within the Design-Bench under protocol 2.
We fix some dimensions and conduct experiments using the bottom 128 samples from the dataset as the $\tilde{\mathbf{x}}$, with $T=250$ and $Q=10$.

\begin{table}[tbp]
    \centering
    \caption{Unnormalized score for Design-Bench.}
    \vspace{-8pt}
        \resizebox{0.985\linewidth}{!}{
        \begin{tabular}{lllll}
        \toprule
        \multirow{2}{*}{} & \multicolumn{2}{c}{\textbf{Hopper Controller}} & \multicolumn{2}{c}{\textbf{TF Bind 8}} \\
                     & \textbf{Mean} & \textbf{Maximum} & \textbf{Mean} & \textbf{Maximum} \\
                     \midrule
                    Grad. & \underline{288.250±23.427} & 1443.152±947.196 & 0.647±0.017 & \underline{0.990±0.003} \\
                    IOM & 271.918±5.893 & 763.319±129.513 & 0.504±0.023 & 0.975±0.012 \\
                    COMs & 285.123±20.812 & 1139.205±712.206 & 0.607±0.017 & 0.984±0.010 \\
                    REM$_p$ & 275.562±18.869 & 1341.529±974.917 &  0.656±0.027 & 0.987±0.008 \\
                    REM$_n$ & 280.942±7.496 & 802.756±116.409 & 0.600±0.028 & 0.973±0.010 \\
                    \midrule
                    \textbf{ROMO}$_p$ & 283.954±11.078 & \underline{1551.855±920.410} & \underline{0.668±0.034} & 0.979±0.013 \\
                    \textbf{ROMO}$_n$ & \textbf{291.905±22.592} & \textbf{1657.105±877.192} & \textbf{0.691±0.032} & \textbf{0.991±0.002} \\
                    \midrule
                    $\tilde{\mathbf{x}}$ & 143.770 & 208.059 & 0.077 & 0.097 \\
                    \bottomrule
        \end{tabular}
        }
    \label{tab:design-bench}
    \vspace{-8pt}
\end{table}

Table~\ref{tab:design-bench} summarizes the main experimental results.
Although all compared methods achieved relatively poor scores since the benchmark was not designed for the CoMBO scenario, we can still observe that ROMO has the best overall performance, with Gradient Ascent closely following and demonstrating significantly better performance than the other methods.\enlargethispage{30pt}

\section{Related Work}
\minisection{Data-driven Model-based Optimization}
Primarily two categories of methods have been investigated for offline MBO.
Inverse generative methods typically utilize a one-to-many inverse mapping to map a target score to ideal designs.
\citet{CbAS, AF-CbAS, PG-VAE} utilize variational autoencoders \cite{vae} to model the design space and well-behave under offline scenarios. \citet{MINs} propose to use a GAN-style \cite{GAN} generator to parameterize the inverse mapping, condition it on target values and involve constraints to search the optimal solutions. \citet{DDOM} use diffusion models
\cite{DDPM, DDIM} to generate the designs manifold, and utilize a conditional diffusion model \cite{dhariwal2021diffusion, ho2021classifierfree} to parameterize the inverse mapping. \citet{bonet} models the optimization process as a sequential decision-making process and use a transformer to generate designs.

The forward surrogate is another way to approach offline model-based optimization problems.
The learned forward mapping serves as a surrogate model of the inaccessible target black-box function.
Therefore, the derivative of function values w.r.t. inputs is accessible, and gradient-based optimizations can be simply applied on the surrogate to find candidates optima.
Typically, additional constraints to the learned surrogate are used to avoid overestimation.
\citet{COMs} introduce explicit conservatism regularizer; \citet{IOMs} utilize invariant representation learning to avoid finding designs that appear erroneously good implicitly.
The surrogate can also serve as a proxy oracle, thus supporting online black-box optimization methods like Bayesian optimization \cite{BeyesianOpt1, BeyesianOpt2, BeyesianOpt3}. 

\minisection{Neighbour-assisted Algorithms}
Neighbour samples are widely applied to assist the prediction.
\citet{IR1, IR2, IR3} propose deep models that treat a target sample $\textbf{x}_t$ as a query, and then retrieve the neighbor and aggregate features to assist the inference of the target sample. \citet{rim} further proposes to utilize the label of retrieved neighbor and consider the interactions between samples.\vspace{-6pt}\enlargethispage{30pt}

\section{Conclusion}
We discuss a novel CoMBO setting, which focuses on optimizing mediocre designs under constraints.
CoMBO is suitable for more application scenarios, but requires better generalization ability and prediction fidelity of the surrogate model.
To this end, we propose \emph{retrieval-enhanced offline model-based optimization} (ROMO), a simple but effective method for forward surrogate learning.
ROMO ensembles a retrieval-enhanced model providing reliable reference predictions, and a surrogate model helping offer effective gradients.
It achieves outstanding performance on a suite of modified tasks in Design-Bench, as well as a real-world load balancing task.
To reveal why ROMO performs better than prior work in the CoMBO setting.
We also provide a suite of analyses with a Hartmann function.

Despite ROMO's good performance in CoMBO, there is still room for improvement.
For future work, we plan to explore the utilization of manifold modeling, and thus map the raw input space to a new manifold, which is easier to optimize with gradient-based methods.
The retrieval process may also be more accurate and efficient in the newly modeled manifold.
Another interesting research direction is to explore whether there are more sophisticated methods than naive gradient ascent to leverage a derivable forward surrogate model and achieve optimization goals.

\bibliographystyle{ACM-Reference-Format}
\bibliography{reference}

\newpage

\appendix

\section{Notations}
Table~\ref{tab:notation} lists important notations and corresponding descriptions.
\begin{table}[htbp]
    \centering
    \caption{Important notations used in this paper.}
    \vspace{-8pt}
    \resizebox{0.985\linewidth}{!}{
        \begin{tabular}{cl}
            \toprule
            \textbf{Notation} & \textbf{Description} \\
            \midrule
            $\mathbf{x}, y$ & Designs and their corresponding scores. \\
            $\mathcal{X}, \tilde{\mathcal{X}}$ & Full design space and its subspace.\\
            $\mathcal{D}$ & Offline dataset.\\
            $\mathcal{D}_{train}, \mathcal{D}_{valid}, \mathcal{D}_{test}$ & Disjoint training set, validation set, testing set. \\
            $f, g, \hat{f}, \hat{g}$ & A function and its approximator. \\
            $\theta, \phi, \mathbf{A}$ & Parameters of neural network. \\
            $\alpha, \beta, \gamma, \lambda, \eta, \tau$ & Hyperparmeters. \\
            $\mathbf{x}^{opt}, \mathbf{x}^{con}$ & Dimensions to be optimized and fixed. \\
            $\mathcal{R}(\cdot)$ & Retrieval candidates. \\
            $\mu(\cdot)$ & Aggregation learning methods. \\
            $\psi(\cdot, \cdot)$ & Similarity function. \\
            $Q$ & Query budgets. \\
            \bottomrule
        \end{tabular}
    }
    \label{tab:notation}
\end{table}

\section{Additional Dataset Details}
\subsection{Hartmann (3D)}
The employed Hartmann (3D) function has the following formulation:
$$ f(\mathbf{x}) = \sum_{i=1}^4 \alpha_i \exp\left(-\sum_{j=1}^3 \mathbf{A}_{ij}\left(x_j-\mathbf{P}_{ij}\right)^2\right),\text{ where}$$
$$ \alpha = \left(1.0, 1.2, 3.0, 3.2\right)^\top,$$
$$
\mathbf{A} = 
    \begin{pmatrix}
    3.0 & 10 & 30 \\
    0.1 & 10 & 35 \\
    3.0 & 10 & 30 \\
    0.1 & 10 & 35 \\
    \end{pmatrix},
$$
$$
\mathbf{P} = 10^{-4}
    \begin{pmatrix}
    3689 & 1170 & 2673 \\
    4699 & 4387 & 7470 \\
    1091 & 8732 & 5547 \\
    381 & 5743 & 8828 \\
    \end{pmatrix}.
$$
The function is evaluated on the hypercube $x_i\in(0,1)$, for all $i=1,2,3$, and has 4 local maxima, and 1 global maxima $f(\mathbf{x}^*)=3.86278$ at $\mathbf{x}^*=(0.114614, 0.555649, 0.852547)$.

\subsection{CIO Load Balancing}
The CIO dataset contains a total of 138 continuous features in $\mathbf{x}$, with 36 dimensions of which represent the CIO parameters that are intended to be modified for the task, while the remaining dimensions represent the physical configuration information of the base stations.

\subsection{Design-Bench}
\begin{figure}[tbp]
    \centering
    \includegraphics[width=0.985\linewidth]{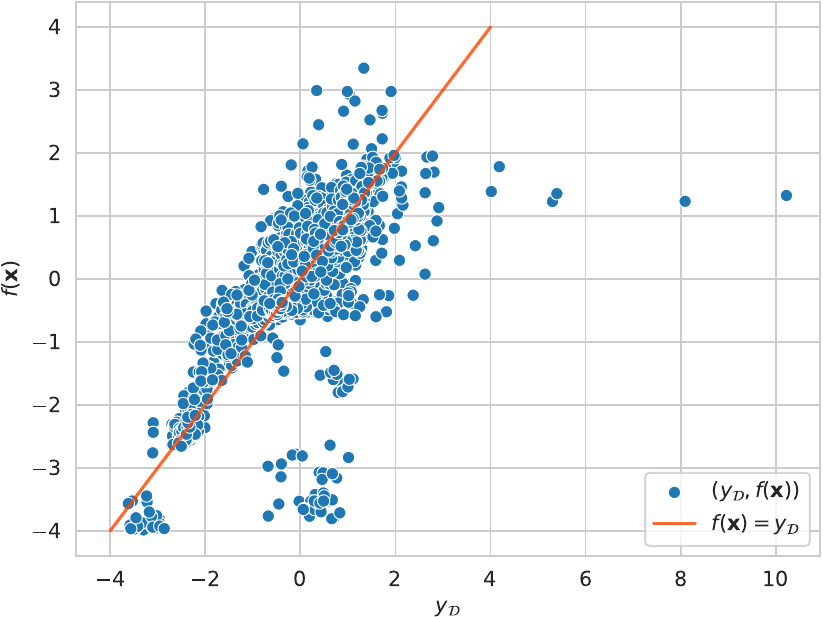}
    \caption{The scatter plot corresponding to the $(y_{\mathcal{D}}, f(\mathbf{x}))$ pairs in the HopperController dataset $\{(\mathbf{x},y_\mathcal{D})\}$, $f$ is the oracle function. The orange line represents the ideal position of the scatter plot. For data points in high-score regions, the oracle still incorrectly assigns mediocre scores.}
    \label{fig:hopper-scatter}
\end{figure}
There is an issue for the Hopper Controller task of Design-Bench as mentioned in \citet{DDOM} and \citet{bonet}. The scores for points in the offline dataset are inconsistent with the predicted scores for the same points given by the oracle (as shown in Figure~\ref{fig:hopper-scatter}). We found that this discrepancy is due to the offline dataset incorrectly using the oracle prepared for predicting normalized features to make predictions on unnormalized samples, thus resulting in a mismatch. To address this issue, we relabeled the offline dataset for Hopper Controller, and using a correct oracle.

\begin{figure}[tbp]
    \centering
    \includegraphics[width=0.985\linewidth]{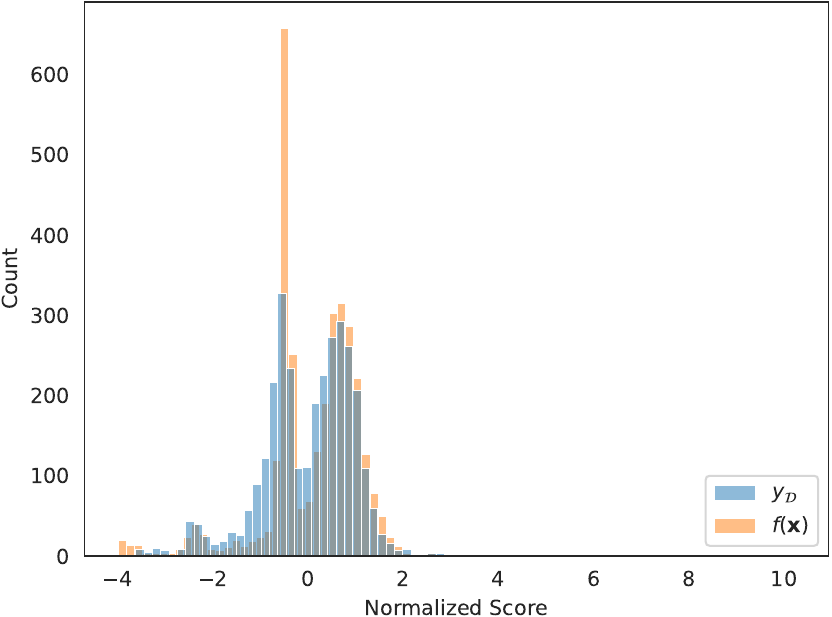}
    \caption{The frequency histogram of the HopperController offline dataset after relabeling. The blue histogram represents the normalized samples after relabeling, while the yellow histogram represents the predicted scores given by a correct oracle for these points. The distribution is highly skewed towards low function values.}
    \vspace{-8pt}
    \label{fig:hopper-hist}
\end{figure}
It is worth noting that even after manually relabeling the HopperController dataset and using the correct oracle for subsequent evaluations, the offline dataset still exhibits the issue of sample points being concentrated in low-score regions, as shown in Figure~\ref{fig:hopper-hist}. Some work has chosen not to use the HopperController dataset due to this reason. However, our argument is that, when relabeling is taken and an correct oracle is utilized, the characteristic of the dataset being concentrated in low-score regions can be considered as an inherent underlying property of the dataset itself and does not need to be treated differently. Moreover, HopperController, which has a disign size of $(3200, 5126)$ is a valuable task instance where the size of offline dataset is small but the dimensions of design space is exceptionally high. Therefore, we still choose to present experimental results based on HopperController in this paper.

\begin{table*}[htbp]
    \centering
    \caption{Design-Bench Experimental Setting Details.}
    \vspace{-8pt}
        \begin{tabular}{llllll}
            \toprule
            \textbf{Task Name} & \textbf{Oracle} & \textbf{Dataset Size} & \textbf{Opt. Dimensions} & \textbf{Task Dimensions} & \textbf{Task Type}\\
            \midrule
            HopperController & Exact & 3200 & 5100 & 5126 & Continuous\\
            TFBind8 & Exact & 32898 & 7 & 8 & Discrete \\
            \bottomrule
        \end{tabular}
    \label{tab:design-bench-xopt}
\end{table*}

To construct a CoMBO setting, we manually fix a portion of each task's last $\tilde{d}$ dimensions, as detailed in Table~\ref{tab:design-bench-xopt}. For a discrete task, we map the one-hot features representing categories to logits and treat them as a continuous task in the model.

\section{Additional Experimental Details}
\begin{table*}[tbp]
    \centering
    \vspace{10pt}
    \caption{Complete Unnormalized Score for the Hartmann Experiment.}
    \vspace{-5pt}
        \begin{tabular}{llllll}
            \toprule
             & \textbf{Mean} & \textbf{100\%} & \textbf{90\%} & \textbf{80\%} & \textbf{50\%} \\
            \midrule
            Grad. & 0.995±0.094 & 2.592±0.199 & 1.915±0.115 & 1.639±0.131 & 0.875±0.137 \\
            IOM & 0.749±0.202 & 2.802±0.310 & 1.650±0.443 & 1.231±0.345 & 0.565±0.267 \\
            COMs & 1.127±0.069 & 2.683±0.188 & 2.097±0.161 & 1.840±0.120 & 1.033±0.065 \\
            REM$_p$ & 0.738±0.486 & 2.244±1.282 & 1.722±1.169 & 1.435±1.025 & 0.547±0.559 \\
            REM$_n$ & 1.135±0.136 & 3.590±0.092 & 3.045±0.107 & 2.521±0.365 & 0.522±0.254 \\
            \midrule
            \textbf{ROMO}$_p$ & \underline{1.758±0.117} & \underline{3.600±0.325} & \underline{3.274±0.242} & \underline{2.846±0.165} & \underline{2.148±0.265} \\
            \textbf{ROMO}$_n$ & \textbf{1.823}±\textbf{0.086} & \textbf{3.721}±\textbf{0.136} & \textbf{3.369}±\textbf{0.165} & \textbf{2.939}±\textbf{0.084} & \textbf{2.294}±\textbf{0.171} \\
            \midrule
            $\tilde{\mathbf{x}}$ & 0.120 & 0.281 & 0.243 & 0.208 & 0.108 \\
            \bottomrule
        \end{tabular}
    \label{tab:hartmann-detail}
\end{table*}

\begin{table*}[tbp]
    \centering
    \vspace{10pt}
    \caption{Complete Normalized Score for the CIO Experiment.}
    \vspace{-5pt}
        \begin{tabular}{llllll}
            \toprule
             & \textbf{Mean} & \textbf{100\%} & \textbf{90\%} & \textbf{80\%} & \textbf{50\%} \\
            \midrule
            Grad. & 93.151±1.763 & \textbf{101.019}±\textbf{1.667} & \textbf{97.795}±\textbf{1.130} & \textbf{96.427}±\textbf{1.400} & 93.459±1.870 \\
            IOM & 92.021±1.410 & 98.287±1.510 & 95.128±1.292 & 94.081±1.364 & 92.052±1.284 \\
            COMs & 92.199±0.434 & 98.642±1.234 & 95.258±0.541 & 94.240±0.561 & 92.265±0.456 \\
            REM$_p$ & 92.760±1.711 & 99.852±1.288 & 96.992±1.432 & 95.889±1.539 & 93.246±1.684 \\
            REM$_n$ & 93.079±1.660 & \underline{99.996±1.097} & 97.131±0.930 & 95.870±1.232 & 93.399±1.603 \\
            \midrule
            \textbf{ROMO}$_p$ & \underline{93.593±0.684} & 99.084±0.948 & 96.597±0.893 & 95.722±0.780 & \underline{93.868±0.711} \\
            \textbf{ROMO}$_n$ & \textbf{94.425}±\textbf{1.087} & 99.415±0.933 & \underline{97.242±0.928} & \underline{96.328±0.938} & \textbf{94.586}±\textbf{0.996} \\
            \midrule
            $\tilde{\mathbf{x}}$ & 82.016 & 83.474 & 83.212 & 82.814 & 82.010 \\
            \bottomrule
        \end{tabular}
    \label{tab:cio-detail}
\end{table*}

\begin{table*}[tbp]
    \centering
    \vspace{10pt}
    \caption{Complete Unnormalized Score for Hopper Controller.}
    \vspace{-5pt}
        \begin{tabular}{llllll}
            \toprule
             & \textbf{Mean} & \textbf{100\%} & \textbf{90\%} & \textbf{80\%} & \textbf{50\%} \\
            \midrule
            Grad. & \underline{288.250±23.427} & 1443.152±947.196 & \underline{528.438±29.579} & \textbf{466.070±39.857} & 212.336±1.376 \\
            IOM & 271.918±5.893 & 763.319±129.513 & 505.526±8.046 & 457.762±6.307 & 208.552±2.833 \\
            COMs & 285.123±20.812 & 1139.205±712.206 & \textbf{533.472±23.730} & 460.507±25.873 & 209.791±1.576 \\
            REM$_p$ & 275.562±18.869 & 1341.529±974.917 & 507.088±17.098 & 449.484±41.567 & 205.249±1.261 \\
            REM$_n$ & 280.942±7.496 & 802.756±116.409 & 524.632±13.848 & 419.578±53.621 & \textbf{222.981±4.295} \\
            \midrule
            \textbf{ROMO}$_p$ & 283.954±11.078 & \underline{1551.855±920.410} & 526.641±19.816 & 456.985±34.789 & 210.301±4.903 \\
            \textbf{ROMO}$_n$ & \textbf{291.905±22.592} & \textbf{1657.105±877.192} & 519.991±16.203 & \underline{463.453±24.472} & \underline{215.562±7.625} \\
            \midrule
            $\tilde{\mathbf{x}}$ & 143.770 & 208.059 & 203.723 & 198.600 & 168.118 \\
            \bottomrule
        \end{tabular}
    \label{tab:hopper-detail}
\end{table*}

\begin{table*}[tbp]
    \centering
    \vspace{10pt}
    \caption{Complete Unnormalized Score for TF Bind 8.}
    \vspace{-5pt}
        \begin{tabular}{llllll}
            \toprule
             & \textbf{Mean} & \textbf{100\%} & \textbf{90\%} & \textbf{80\%} & \textbf{50\%} \\
            \midrule
            Grad. & 0.647±0.017 & \underline{0.990±0.003} & 0.900±0.025 & 0.802±0.033 & 0.614±0.016 \\
            IOM & 0.504±0.023 & 0.975±0.012 & 0.729±0.056 & 0.615±0.042 & 0.466±0.026 \\
            COMs & 0.607±0.017 & 0.984±0.010 & 0.878±0.020 & 0.811±0.017 & 0.572±0.035 \\
            REM$_p$ & 0.656±0.027 & 0.987±0.008 & \underline{0.910±0.031} & 0.844±0.031 & \underline{0.648±0.055} \\
            REM$_n$ & 0.600±0.028 & 0.973±0.010 & 0.858±0.032 & 0.749±0.033 & 0.589±0.037 \\
            \midrule
            \textbf{ROMO}$_p$ & \underline{0.668±0.034} & 0.979±0.013 & 0.905±0.041 & \underline{0.850±0.062} & 0.646±0.055 \\
            \textbf{ROMO}$_n$ & \textbf{0.691±0.032} & \textbf{0.991±0.002} & \textbf{0.946±0.010} & \textbf{0.891±0.024} & \textbf{0.702±0.056} \\
            \midrule
            $\tilde{\mathbf{x}}$ & 0.077 & 0.097 & 0.095 & 0.092 & 0.084 \\
            \bottomrule
        \end{tabular}
    \label{tab:tfbind8-detail}
\end{table*}

The complete normalized performance for the experiment of Hartmann (3D) test function is detailed in Table~\ref{tab:hartmann-detail};
the complete normalized performance for the experiment of CIO load balancing is detailed in Table~\ref{tab:cio-detail};
The complete unnormalized performance for the experiments of HopperController and TF Bind 8 are detailed in Table~\ref{tab:hopper-detail} and Table~\ref{tab:tfbind8-detail}, respectively.

\subsection{Implementation Details}
\minisection{Similarity Function}
The similarity function with a RBF kernel product is defined as
\begin{equation}
    \psi_{RBF}(\textbf{x}_i,\textbf{x}_j) = \exp\left(-\frac{\|\textbf{x}_i - \textbf{x}_j\|^2_2}{2\sigma^2}\right),
\end{equation}
where the $\sigma$ is a bandwidth parameter.

The cosine similarity is defined as
\begin{equation}
    \psi_{cos}(\textbf{x}_i,\textbf{x}_j) = \frac{\textbf{x}_i\cdot\textbf{x}_j}{\|\textbf{x}_i\|\cdot\|\textbf{x}_j\|}.
\end{equation}

\minisection{Pseudo Code}
The pseudo code for the training and inference processes of ROMO are provided by Alogrithm~\ref{alg:train} and Algorithm~\ref{alg:test}, respectively.

At the training time, datasets are firstly divided into disjoint training, validation and retrieval pool sets. For each training step, we construct retrieval candidates for current training samples, and then perform dual optimization with the training objective given by Equation~\ref{eq:train-dual}.

In the inference process, mediocre initial designs $\tilde{\mathbf{x}}$ are first selected. We then perform $T$ steps of gradient ascent using the gradient of $\hat{y}$ (output of $\hat{h}_{\theta,\phi}$) w.r.t. $\mathbf{x}^{opt}$ for $\tilde{\mathbf{x}}$. The derived $\mathbf{x}_T$s are treated as the candidate solutions.

\begin{algorithm}[tbp]
\caption{ROMO: Training with Objective of Eq.~\eqref{eq:train-dual}}\label{alg:train}
\begin{algorithmic}[1]
\State Initialize forward model $\hat{f}_\theta$,$\hat{g}_\phi$, dataset $\mathcal{D}_{train}, \mathcal{D}_{valid},\mathcal{D}_{pool}$
\State Construct Lagrangian $\mathcal{L}$ and dual function $\text{g}$ for Eq.\eqref{eq:train-dual}:
\Statex \quad $\mathcal{L}((\theta,\phi);\lambda) = \mathcal{L}_s + \mathcal{L}_a + \lambda(\hat{f}_\theta(\textbf{x}) - \hat{g}_\phi(\textbf{x},\mu(\mathcal{R}(\textbf{x})))-\tau)$
\Statex \quad $\text{g}(\lambda) = \mathcal{L}((\theta^*,\phi^*);\lambda)$, \text{ where} $(\theta^*,\phi^*)=\arg\min_{\theta,\phi}\mathcal{L}(\lambda)$
\State Pick learning rate $\eta$, $\alpha$, training steps $T$
\For{$t=1,\ldots,T$}
    \State Sample $(\textbf{x}, y)\sim \mathcal{D}_{train}$
    \State Retrieve $\mathcal{R}(\textbf{x})\sim\mathcal{D}_{pool}$ for aggregations $\mu(\mathcal{R}(\textbf{x}))$
    \State $\theta \leftarrow \theta - \eta\nabla_\theta\mathcal{L}((\theta,\phi);\lambda)$;\quad$\phi \leftarrow \phi - \eta\nabla_\phi\mathcal{L}((\theta,\phi);\lambda)$
    \State $\lambda \leftarrow \lambda + \alpha\nabla_\lambda\text{g}(\lambda)$
\EndFor
\State \Return $\hat{h}_{\theta,\phi}$
\end{algorithmic}
\end{algorithm}

\begin{algorithm}[tbp]
\caption{ROMO: Finding ${\textbf{x}^{opt}}^{*}$ for a CoMBO Problem}\label{alg:test}
\begin{algorithmic}[1]
\State Initialize dataset $\mathcal{D}_{pool}, \mathcal{D}_{test}$, mediocre threshold $y_{thres}$
\Procedure{Mediocre}{$\mathcal{D}$} \Comment{Find mediocre initial designs.}
    \State $\tilde{\textbf{x}} = \{\textbf{x}|(\textbf{x},y)\in\mathcal{D}_{test}, y<y_{thres}\}$
    \State \Return $\tilde{\textbf{x}}$
\EndProcedure
\Procedure{Optimize}{} \Comment{Find ${\textbf{x}^{opt}}^*$ via gradient ascent.}
    \State Initial optimizer with $\textbf{x}_0 = \text{MEDIOCRE}(\mathcal{D}_{test})$
    \State Pick model $\hat{h}_{\theta,\phi}$, learning rate $\eta$, optimization steps $T$
    \For{$t=1, \ldots, T$}
        \State Retrieve $\mathcal{R}(\textbf{x}_{t-1})\sim\mathcal{D}_{pool}$ for aggregations $\mu(\mathcal{R}(\textbf{x}))$
        \State $\textbf{x}^{opt}_t = \textbf{x}^{opt}_{t-1} + \eta\nabla_{\textbf{x}^{opt}}\mathcal{L}(\textbf{x})|_{\textbf{x}=\textbf{x}_{t-1}}$,
        \Statex \quad \quad \quad where $\mathcal{L}(\textbf{x}) := \hat{h}_{\theta,\phi}(\textbf{x},\mu(\mathcal{R}(\textbf{x})))$
    \EndFor
    \State \Return ${\textbf{x}^{opt}}^* = \textbf{x}^{opt}_T$
\EndProcedure
\end{algorithmic}
\end{algorithm}

\subsection{Training Detials}
We train our model on one RTX 3080 Ti GPU and report results averaged over 5 seeds for all tasks.

In a parameterized aggregation learning process, we take the $\gamma$ of Equation~\eqref{eq:gamma} with $\gamma = 2$ in our experiments.
For the $\lambda$ hyperparameter in Equation~\eqref{eq:solve-ridge}, we take $\lambda = 0.1$ in our experiments.

The $\beta$ used to perform a weighted ensemble for the surrogate forward model and the retrieval-enhanced forward model is taken as $\beta = 0.5$ for all the experiments.

For the $\tau$ used in conservatism regularizer proposed in Equation~\eqref{eq:train-dual}, we adopt similar parameters as in COMs \cite{COMs} for Design-Bench, where $\tau = 0.5$ for continuous tasks (Hopper Controller), and $\tau = 2.0$ for discrete tasks (TF Bind 8). As for the Hartmann task and CIO task, we take $\tau = 0.01$.

\section{Additional Discussions}
\begin{figure}[tbp]
    \centering
    \includegraphics[width=0.985\linewidth]{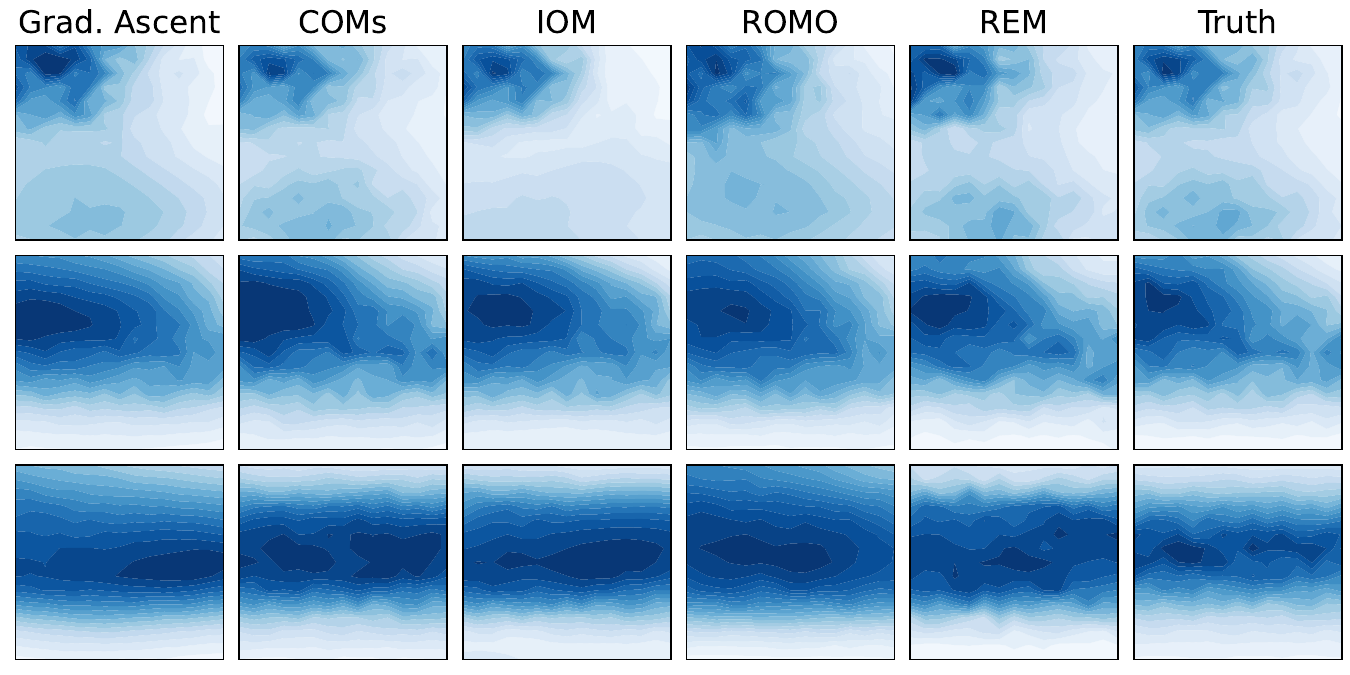}
    \caption{Manifolds Learned by Different Models. The rightmost column is the ground truth manifolds.}
    \label{fig:manifold}
    \vspace{-8pt}
\end{figure}

\minisection{Manifold Learning}
In order to compare the differences between the learned manifolds, and further validate some of the analysis mentioned in Section~\ref{sec:experiments}, we visualize the manifolds as in Figure~\ref{fig:manifold}.

The manifolds learned by COMs exhibit good accuracy. However, they sometimes can indeed lead to over-underestimation, as shown by the manifold in the second row, which may further cause the ``fake local optima''.
IOM's manifolds align with its design intent. It tends to learn more uniformly flat manifolds, which also results in IOM performing averagely in CoMBO scenarios that involve caring about mediocre data points.
With the assistance of retrieved aggregations, REM achieves the best generalization and more accurate manifolds in the entire space. However, our analysis indicates that the additional dimensions introduced by $\mu(\mathcal{R}(\mathbf{x}))$ cause directly optimizing on REM's manifold typically does not yield strong performance.
Although the manifolds learned by ROMO are not the most precise, they do exhibit certain advantages compared to IOM and Gradient Ascent. Additionally, ROMO avoids the over-underestimation issues seen in COMs. As a result, ROMO can achieve excellent performance in Hartmann, as demonstrated in Table~\ref{tab:hartmann}.

\begin{figure}[tbp]
    \centering
    \includegraphics[width=0.985\linewidth]{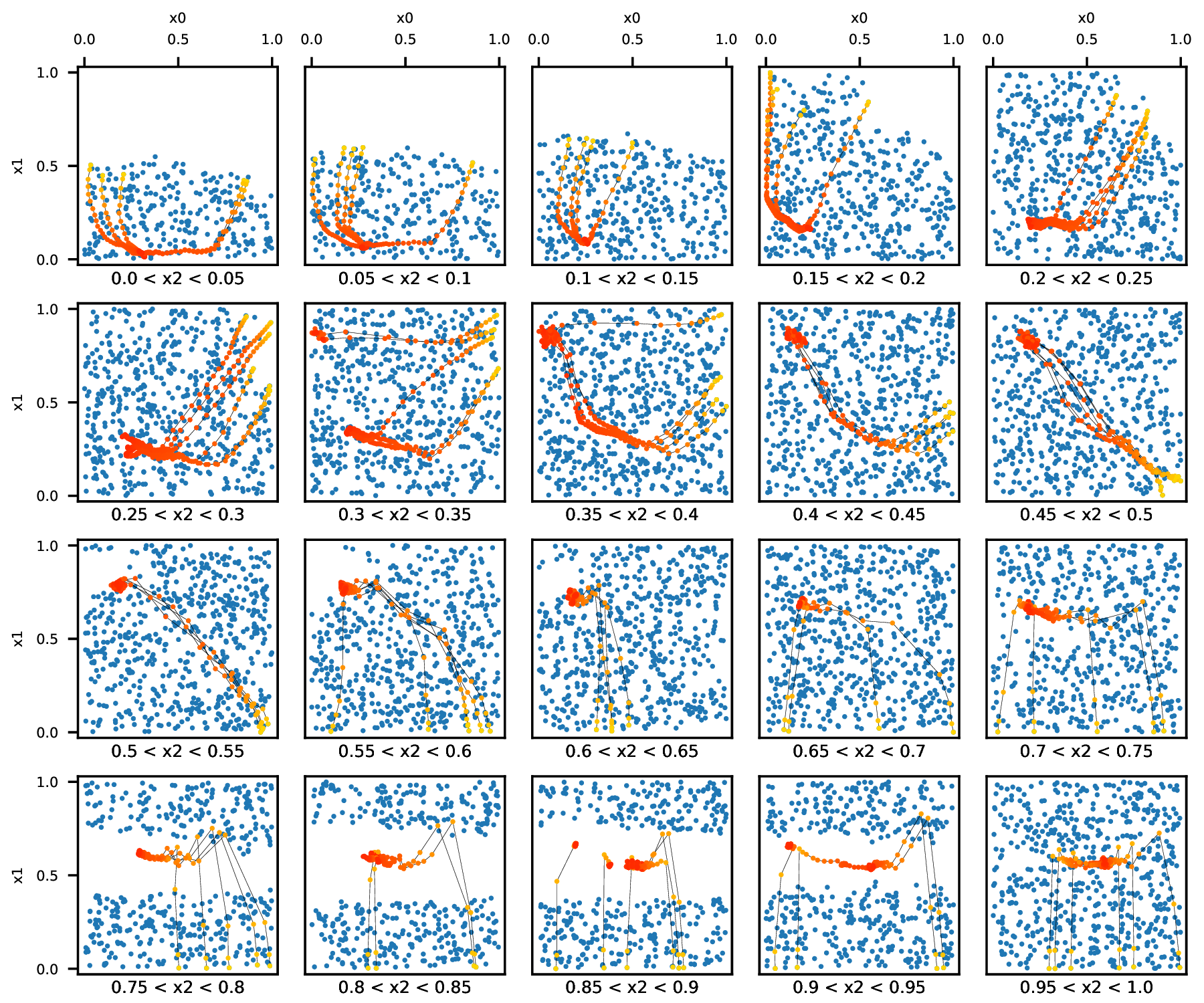}
    \caption{Visualization of offline data support and the final candidate solutions found by ROMO. The blue scatter points represent the offline samples in the dataset, and the red points show where the found candidate solutions locate.}
    \label{fig:generalization}
    \vspace{-5pt}
\end{figure}

\minisection{The Ability to Generalize Beyond the Dataset}
To discuss the ability of ROMO to generalize beyond the offline data support,
we provide a visualization as is shown in Figure~\ref{fig:generalization}.
The blue scatter points illustrate the offline data support.
The forward surrogate learned by ROMO demonstrates good generalization ability during the CoMBO optimization process.
ROMO can generalize outside the data support and search for potential optimal candidates located in the unknown region.

\begin{figure*}[h!]
    \centering
    \captionsetup[subfloat]{labelformat=empty}
    \subfloat[Grad. Learned Manifold]{
        \centering
        \includegraphics[width=0.31\textwidth]{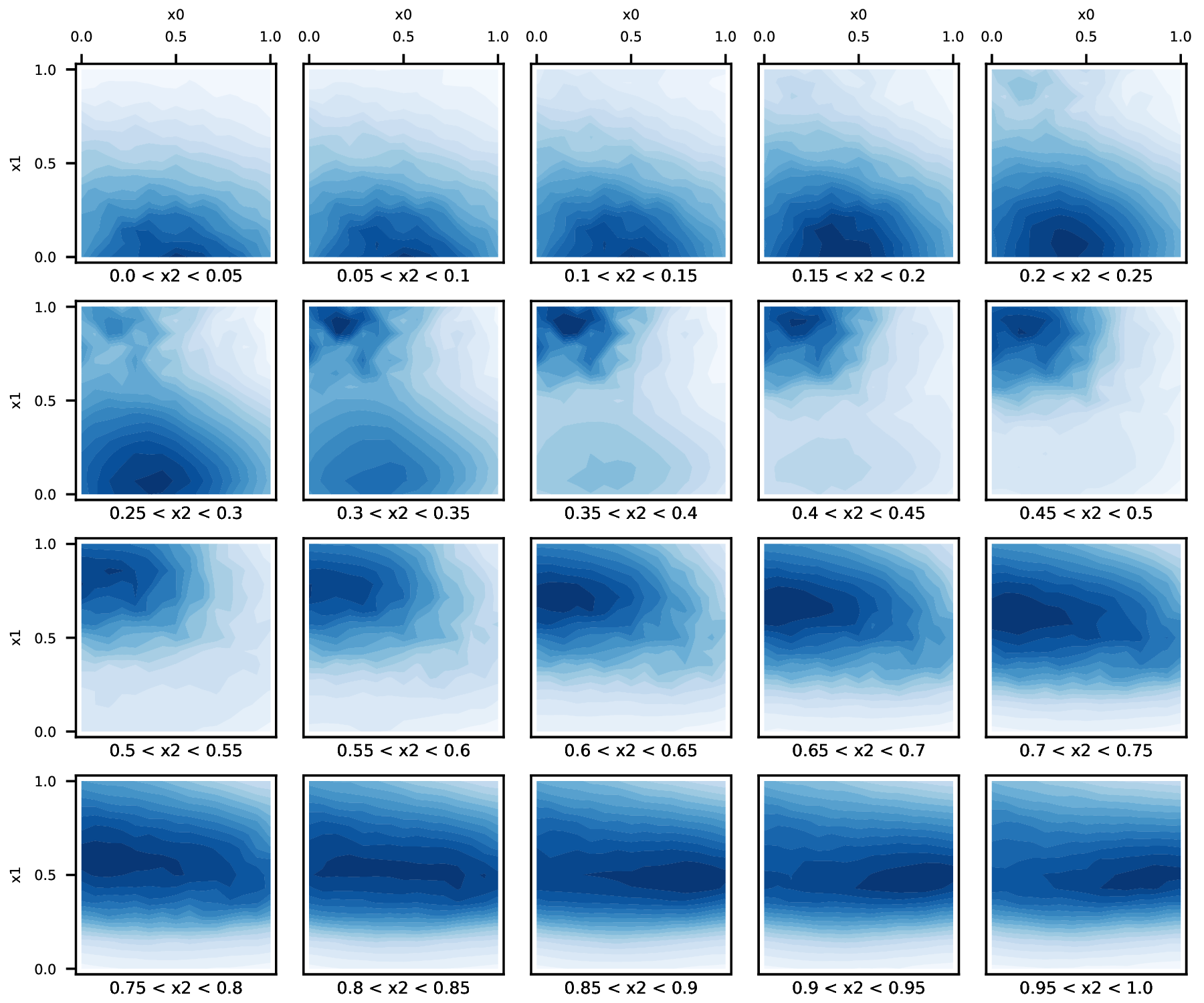}
    }
    \subfloat[Grad. Candidates location]{
        \centering
        \includegraphics[width=0.31\textwidth]{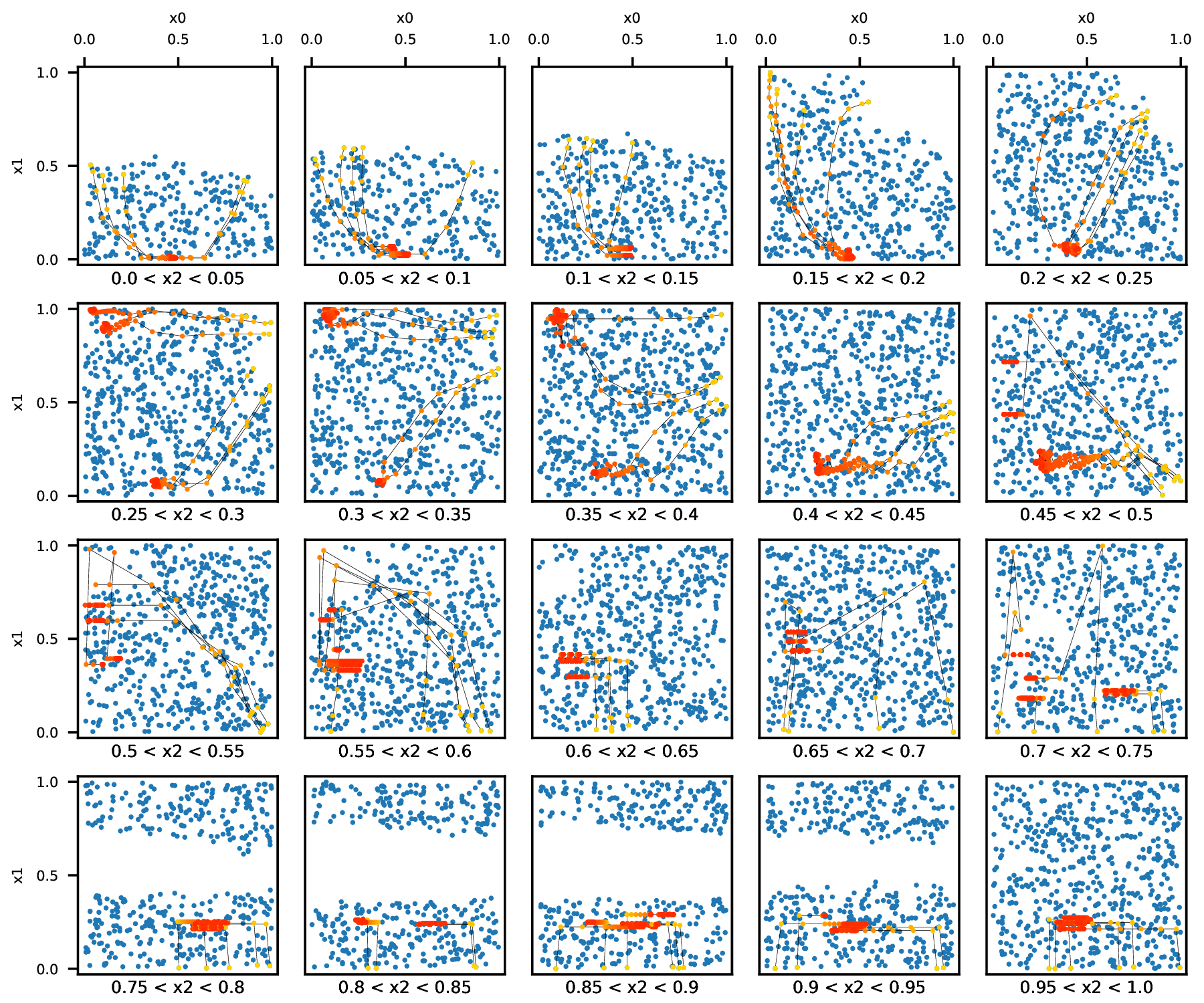}
    }
    \subfloat[Grad. Model Behavior]{
        \centering
        \includegraphics[width=0.31\textwidth]{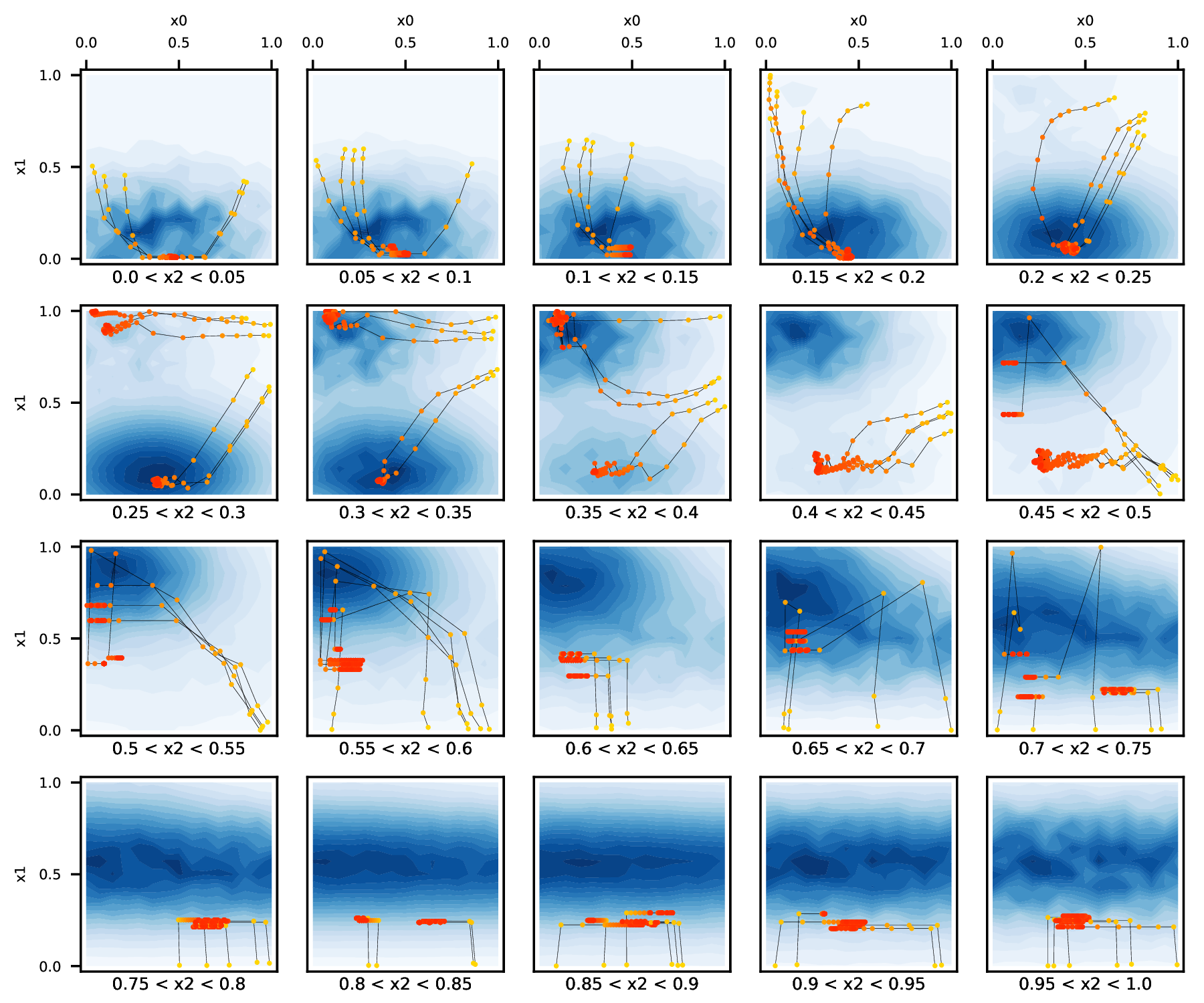}
    }
    \vspace{-10pt}
    \label{fig:grad-all}
\end{figure*}
\begin{figure*}[h!]
    \centering
    \captionsetup[subfloat]{labelformat=empty}
    \subfloat[IOM Learned Manifold]{
        \centering
        \includegraphics[width=0.31\textwidth]{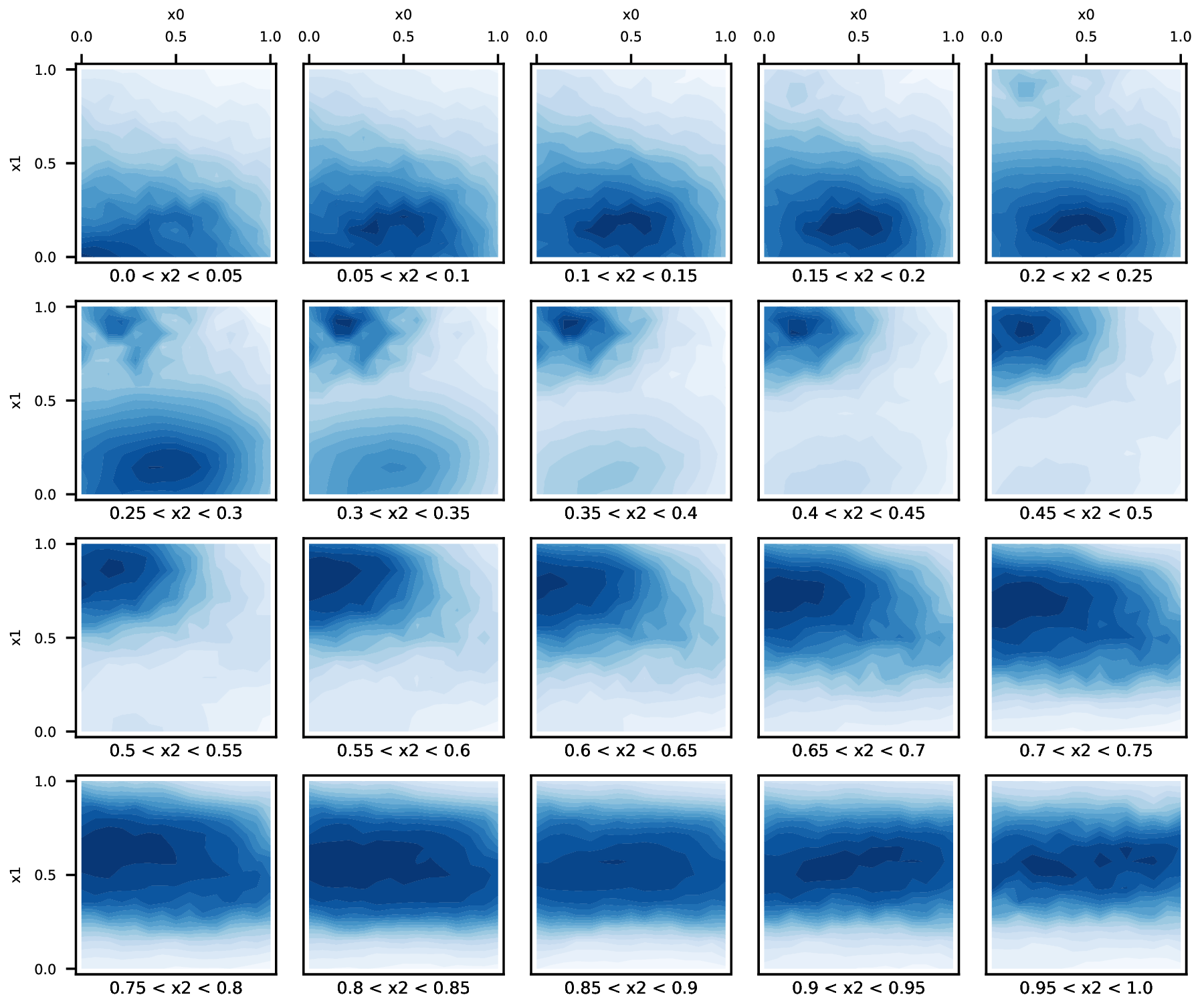}
    }
    \subfloat[IOM Candidates location]{
        \centering
        \includegraphics[width=0.31\textwidth]{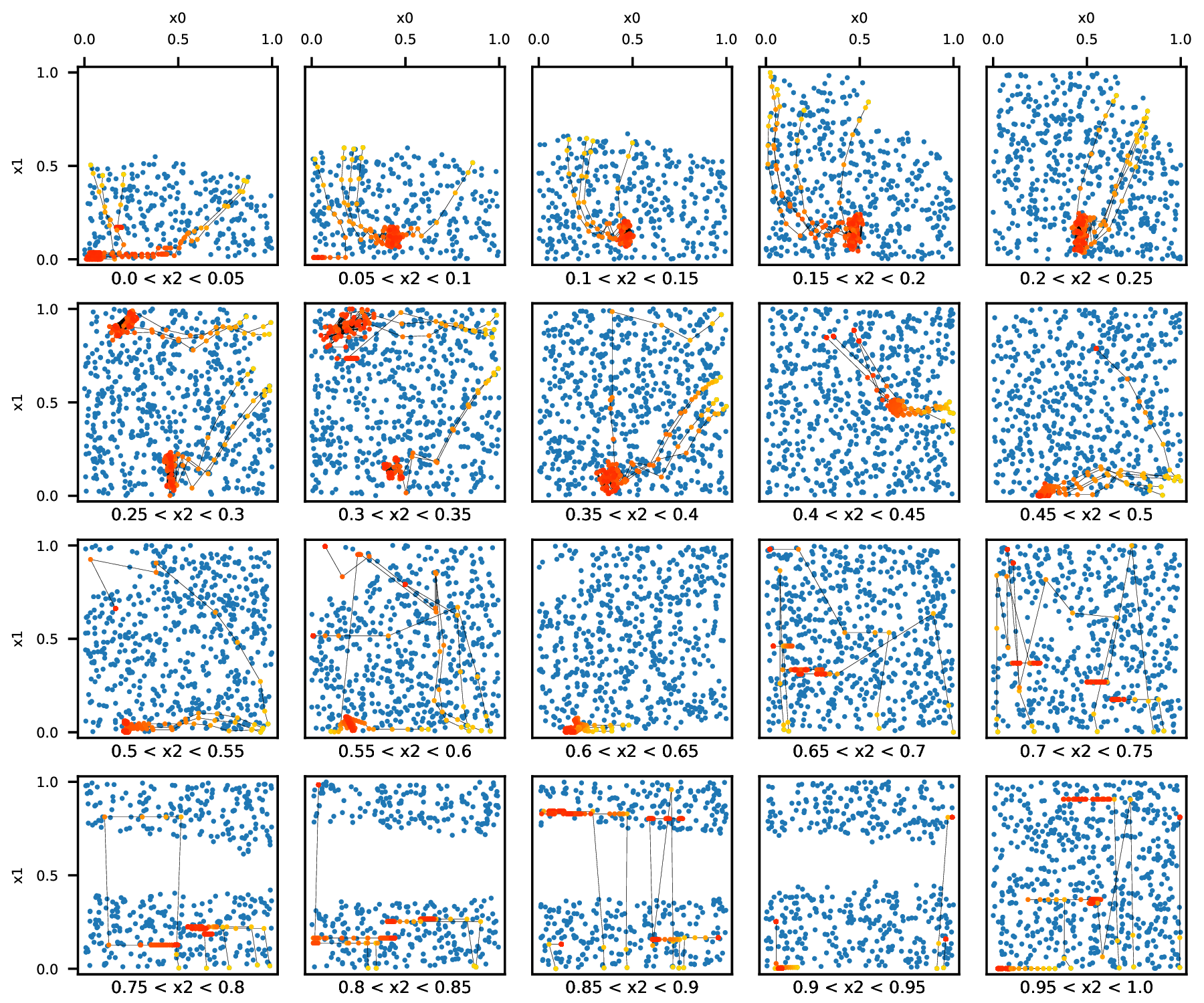}
    }
    \subfloat[IOM Model Behavior]{
        \centering
        \includegraphics[width=0.31\textwidth]{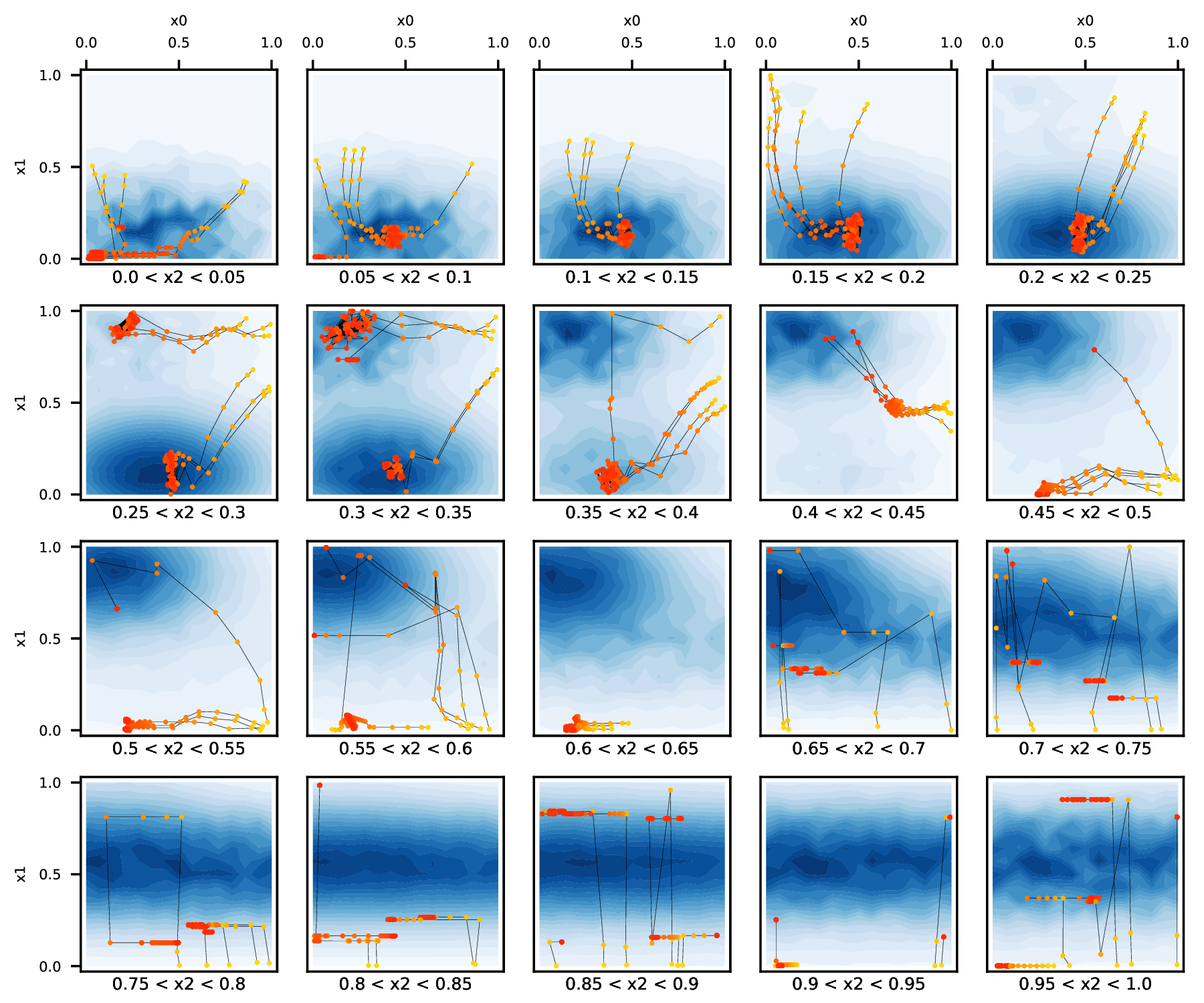}
    }
    \vspace{-10pt}
    \label{fig:grad-all}
\end{figure*}
\begin{figure*}[h!]
    \centering
    \captionsetup[subfloat]{labelformat=empty}
    \subfloat[COMs Learned Manifold]{
        \centering
        \includegraphics[width=0.31\textwidth]{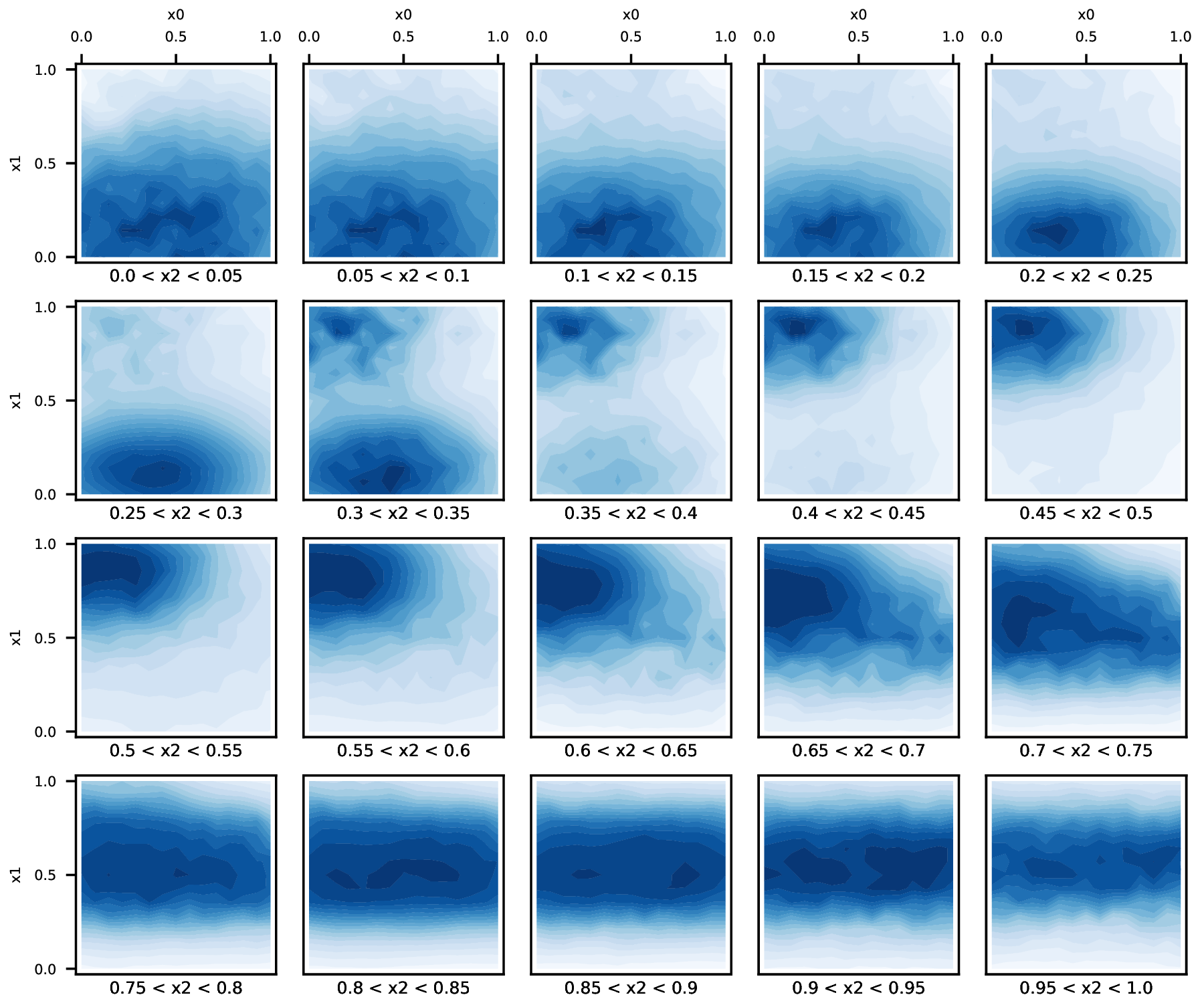}
    }
    \subfloat[COMs Candidates location]{
        \centering
        \includegraphics[width=0.31\textwidth]{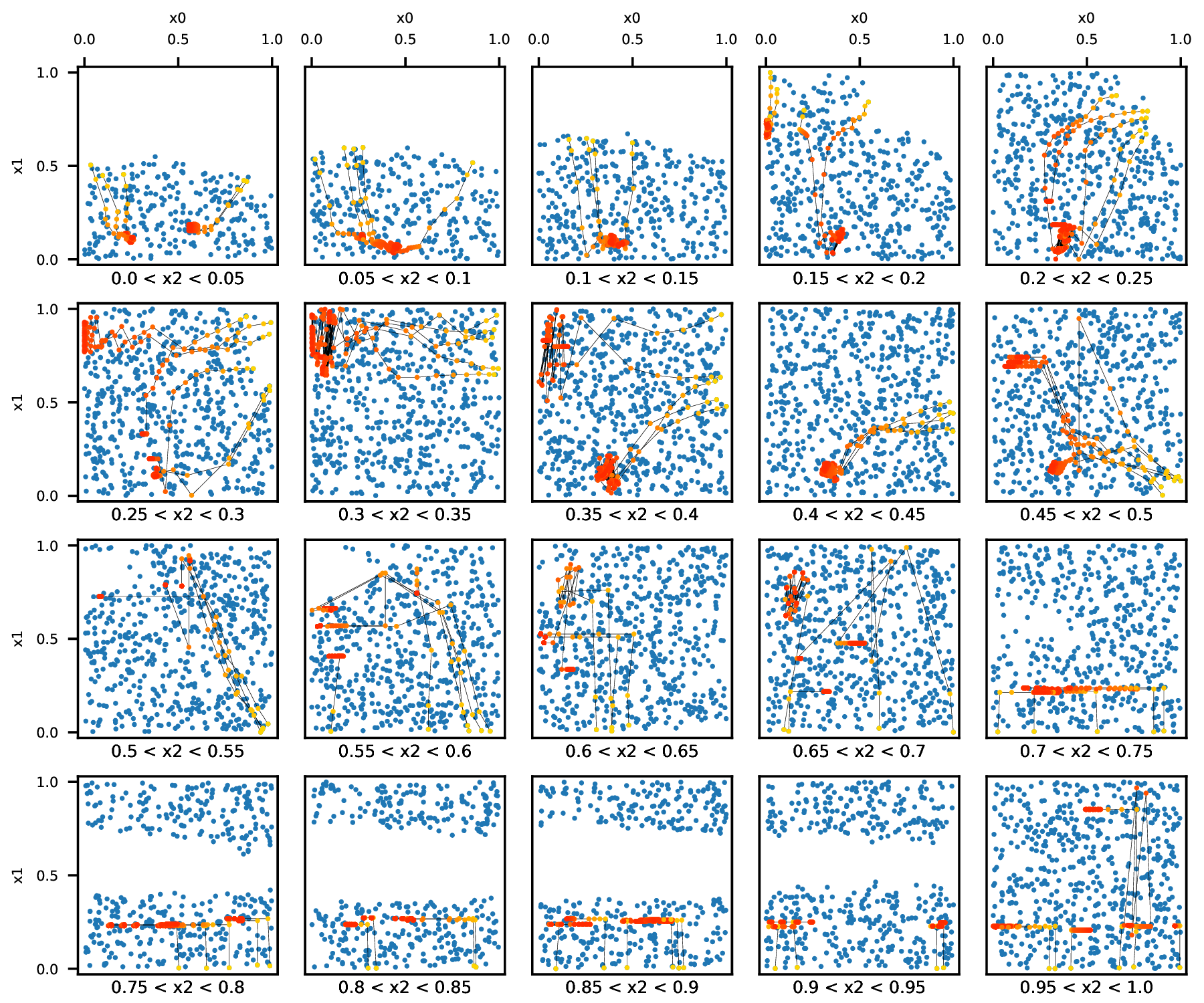}
    }
    \subfloat[COMs Model Behavior]{
        \centering
        \includegraphics[width=0.31\textwidth]{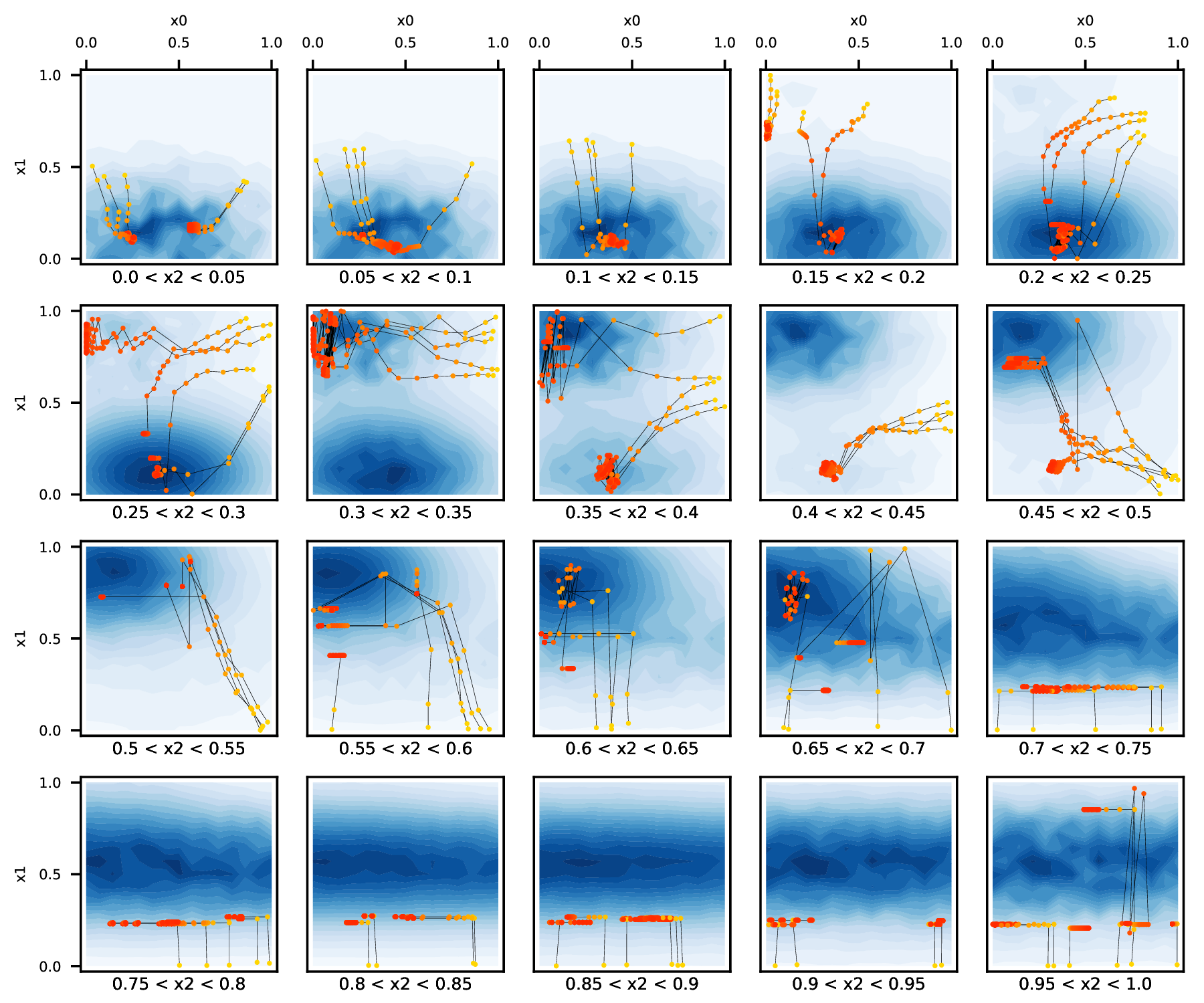}
    }
    \vspace{-10pt}
    \label{fig:grad-all}
\end{figure*}
\begin{figure*}[h!]
    \centering
    \captionsetup[subfloat]{labelformat=empty}
    \subfloat[REM Learned Manifold]{
        \centering
        \includegraphics[width=0.31\textwidth]{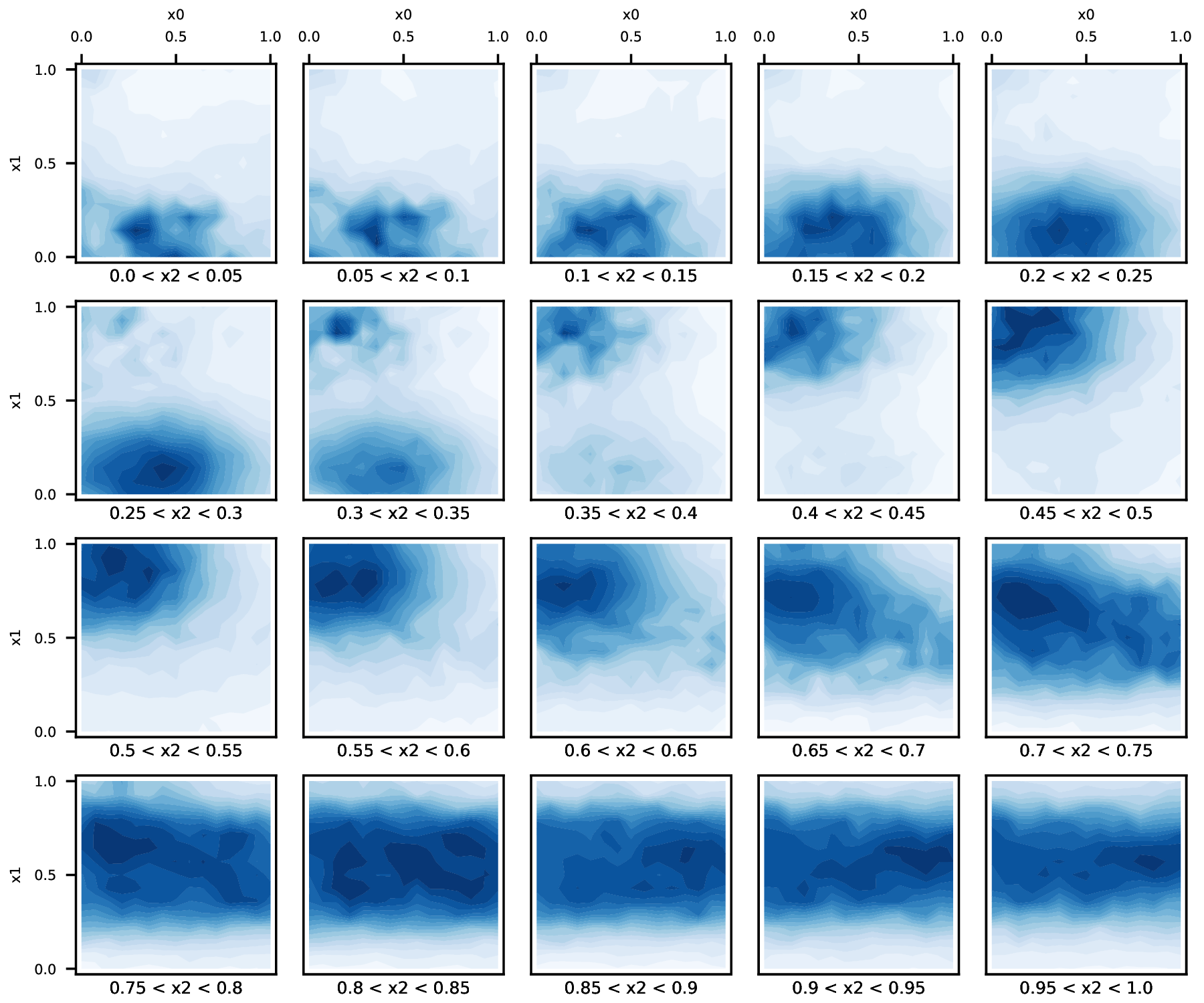}
    }
    \subfloat[REM Candidates location]{
        \centering
        \includegraphics[width=0.31\textwidth]{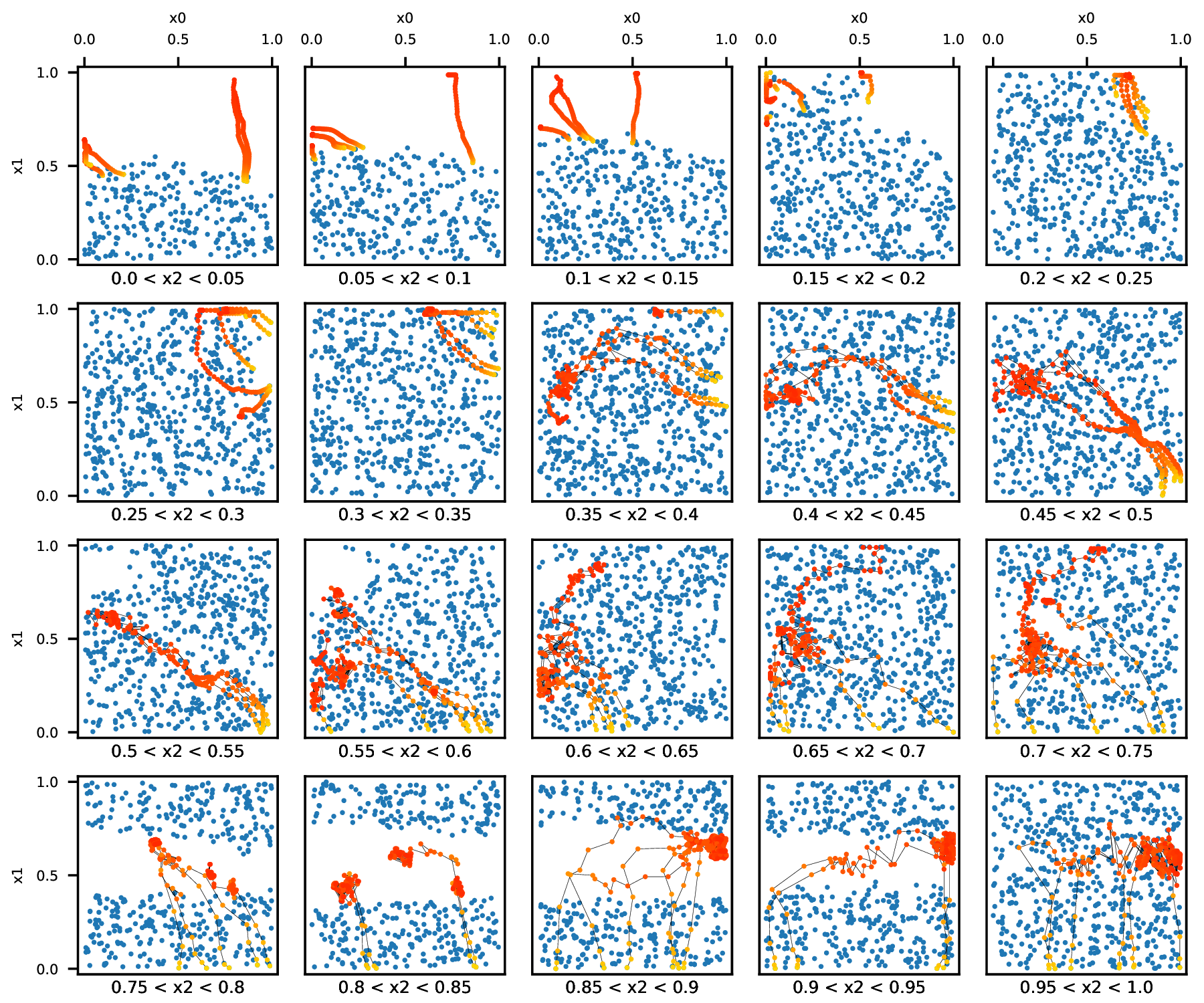}
    }
    \subfloat[REM Model Behavior]{
        \centering
        \includegraphics[width=0.31\textwidth]{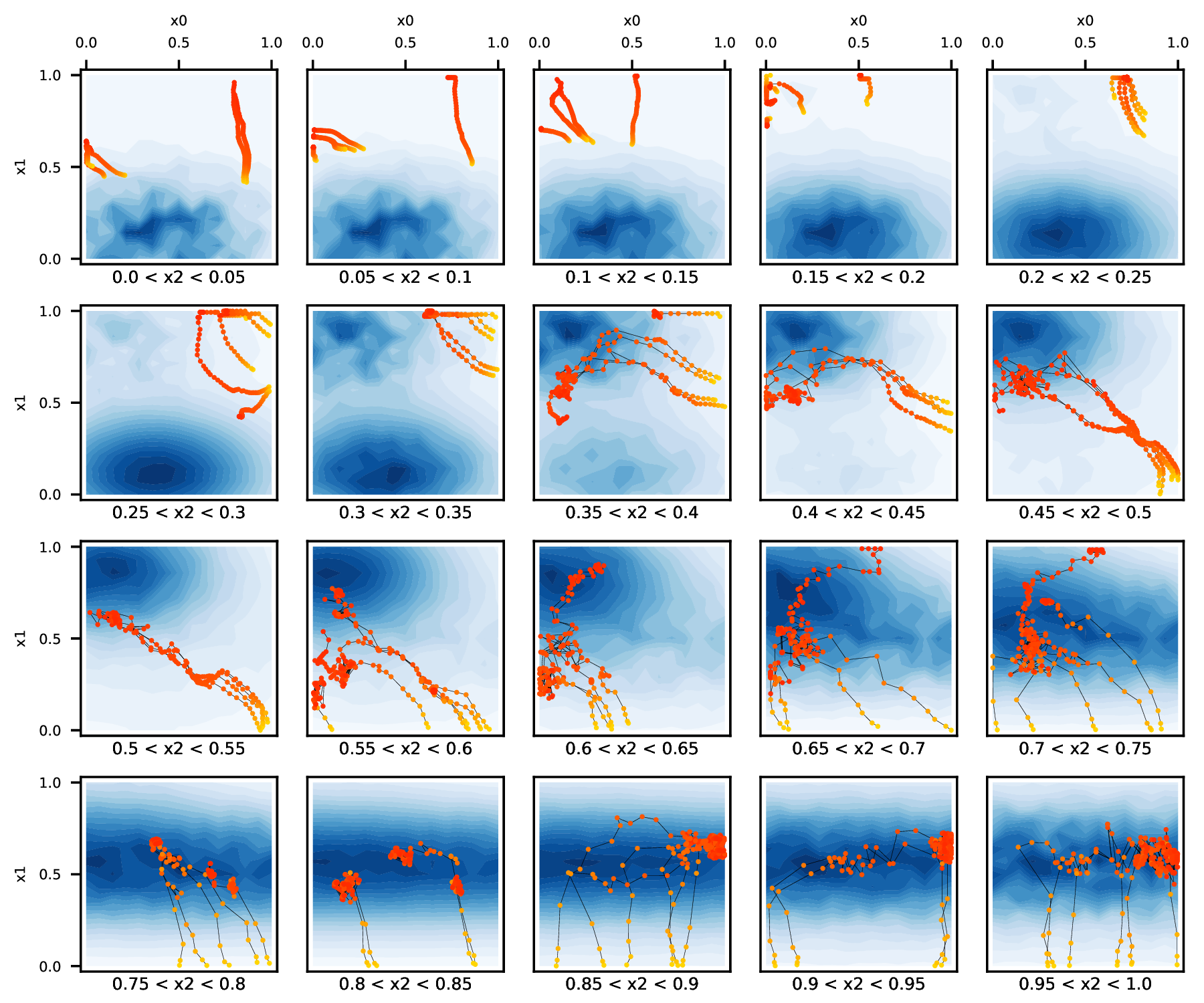}
    }
    \vspace{-10pt}
    \label{fig:grad-all}
\end{figure*}
\begin{figure*}[h!]
    \centering
    \captionsetup[subfloat]{labelformat=empty}
    \subfloat[ROMO Learned Manifold]{
        \centering
        \includegraphics[width=0.31\textwidth]{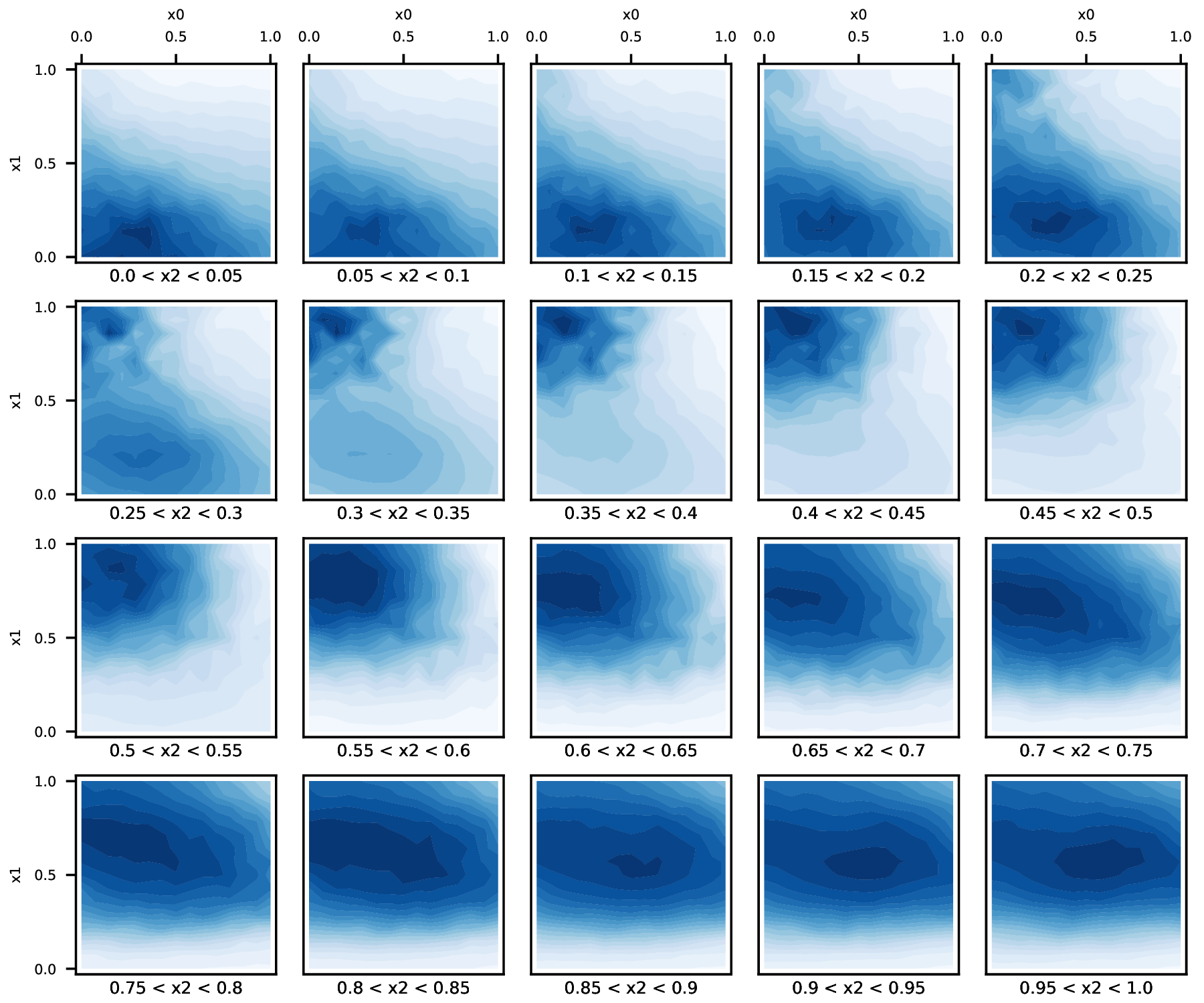}
    }
    \subfloat[ROMO Candidates location]{
        \centering
        \includegraphics[width=0.31\textwidth]{./figs/romo_samples.png}
    }
    \subfloat[ROMO Model Behavior]{
        \centering
        \includegraphics[width=0.31\textwidth]{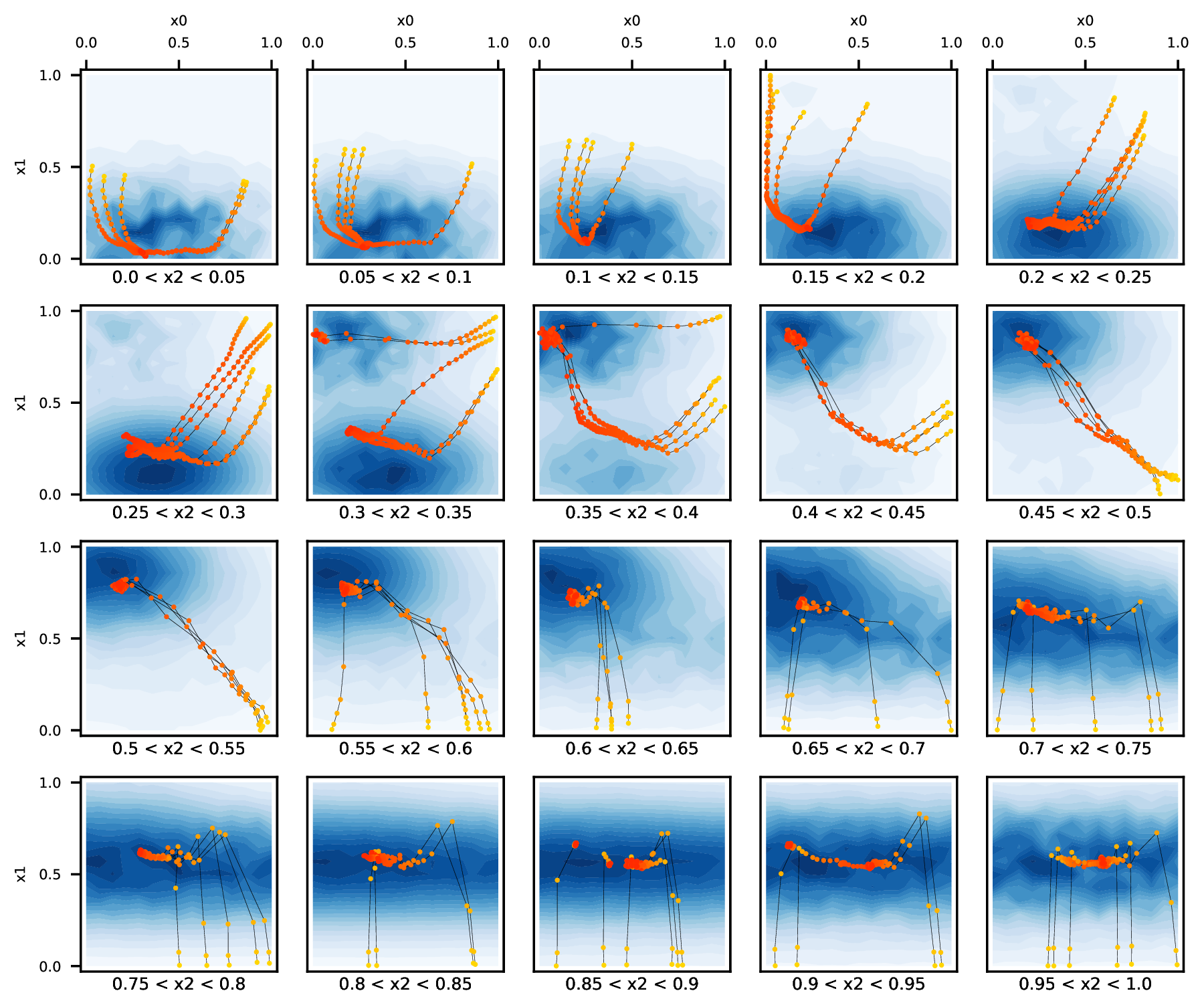}
    }
    \caption{An overview of horizontal and vertical comparisons. Each row corresponds to a CoMBO method, and each column corresponds to a visualization method. All results are obtained with the Hartmann (3D) test function.}
    \label{fig:all}
\end{figure*}

\minisection{Overall Comparison}
In the end, to provide a convenient horizontal and vertical comparison, we organize an overview of (i). learned manifold; (ii). candidates location; (iii). model behavior for each compared method as in Figure~\ref{fig:all}. Each row is for one of the compared methods, while each column shows one type of visualization results.
Due to the use of the kernel density estimation (KDE) method in plotting the manifold, the final presentation of the learned manifold may not faithfully capture the local fluctuations and minor differences in the manifold. Therefore, it may be challenging to discern the direct impact of different manifolds on the CoMBO optimization performance.

Nonetheless, we can also conclude that in most cases, Gradient Ascent with a naive fully connected neural network remains a reasonably balanced and reliable method for CoMBO, even though the forward model fails at generalizing at the edge of the offline data support;

To some extent, IOM can exhibit a degree of generalization in the OOD regions. However, excessively homogeneous manifolds can be disadvantageous for gradient-based optimization in a CoMBO setting;

COMs shows excessive conservatism in the CoMBO setting, and in many instances, it misses opportunities for improvement because it cannot venture outside the data support;

REM, with a retrieval-enhanced model only, provides the most accurate learned manifold that can generalize well outside the data support and provide the opportunity for leaving the support and searching for potential optimal candidates.
However, the impact of the additional dimensions brought by aggregations may cause gradient steps to drift uncontrollably toward an undesirable direction;

Ensembled with a retrieval-enhanced model, ROMO has successfully achieved robust generalization across the entire design space and can discover candidates beyond offline data support.

\end{document}